\documentclass[journal]{IEEEtran}

\usepackage{times}
\usepackage{epsfig}
\usepackage{graphicx}
\usepackage{amsmath}
\usepackage{amssymb}
\usepackage{multirow}
\usepackage{makecell}

\usepackage{subfig}

\usepackage{dblfloatfix} 
\usepackage{booktabs}
\usepackage{siunitx}
\usepackage[pagebackref=true,breaklinks=true,bookmarks=false]{hyperref}
\usepackage[capitalize,nameinlink]{cleveref}

\begin{document}

\title{Deep Learning for Large-Scale Traffic-Sign Detection and Recognition}

\author{Domen Tabernik and Danijel Sko\v{c}aj\\
Faculty of Computer and Information Science, University of Ljubljana\\
Ve\v{c}na pot 113, 1000 Ljubljana\\
{\tt\small \{domen.tabernik,danijel.skocaj\}@fri.uni-lj.si}
}

\maketitle

\begin{abstract}
Automatic detection and recognition of traffic signs plays a crucial role in management of the traffic-sign inventory. It provides accurate and timely way to manage traffic-sign inventory with a minimal human effort. In the computer vision community the recognition and detection of traffic signs is a well-researched problem. A vast majority of existing approaches perform well on traffic signs needed for advanced drivers-assistance and autonomous systems. However, this represents a relatively small number of all traffic signs (around 50 categories out of several hundred) and performance on the remaining set of traffic signs, which are required to eliminate the manual labor in traffic-sign inventory management, remains an open question. In this paper, we address the issue of detecting and recognizing a large number of traffic-sign categories suitable for automating traffic-sign inventory management. We adopt a convolutional neural network (CNN) approach, the Mask R-CNN, to address the full pipeline of detection and recognition with automatic end-to-end learning. We propose several improvements that are evaluated on the detection of traffic signs and result in an improved overall performance. This approach is applied to detection of 200 traffic-sign categories represented in our novel dataset. Results are reported on highly challenging traffic-sign categories that have not yet been considered in previous works. We provide comprehensive analysis of the deep learning method for the detection of traffic signs with large intra-category appearance variation and show below \num{3}{\%} error rates with the proposed approach, which is sufficient for deployment in practical applications of traffic-sign inventory management.
\end{abstract}



\section{Introduction}

\IEEEPARstart{P}{roper} management of traffic-sign inventory is an important task in ensuring safety and efficiency of the traffic flow~\cite{Balali2015,Wang2010}. Most often this task is performed manually. Traffic signs are captured using a vehicle-mounted camera and manual localization and recognition is performed off-line by a human operator to check for consistency with the existing database. However, such manual work can be extremely time-consuming when applied to thousands of kilometers of roads. Automating this task would significantly reduce the amount of manual work and improve safety through quicker detection of damaged or missing traffic signs~\cite{Balali2016}.

\begin{figure}
\includegraphics[width=\columnwidth]{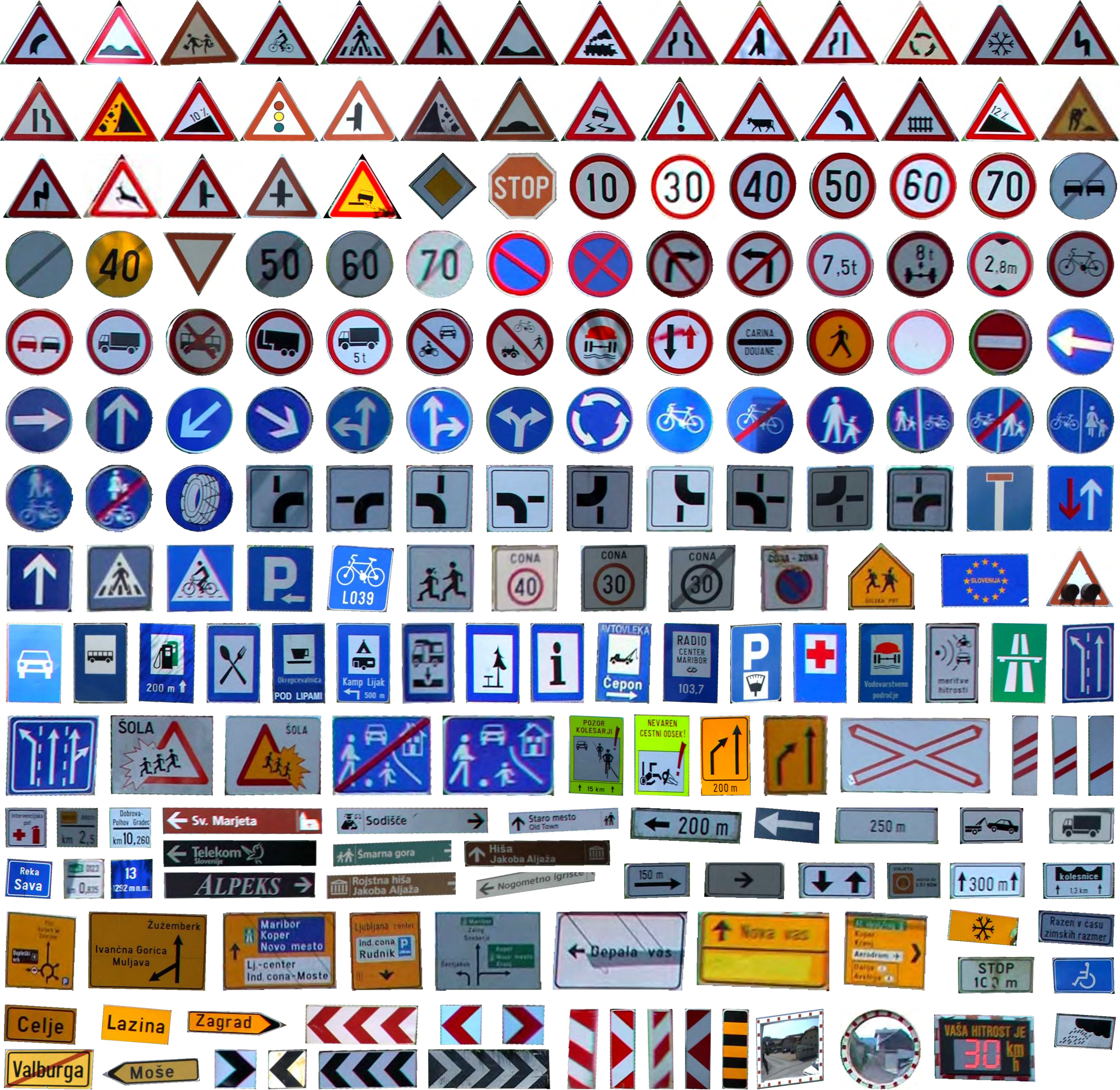}
\caption{The DFG traffic-sign dataset consists of 200 categories including large number of traffic signs with high intra-category appearance variations.\label{fig:dfg-database}}
\end{figure}

A crucial step towards the automation of this task is replacing manual localization and recognition of traffic signs with an automatic detection. In the computer-vision community the problem of traffic-sign recognition has already received a considerable attention~ \cite{LilloCastellano2015,Haloi2015a,Zhu2016}, and excellent detection and recognition algorithms have already been proposed. But these solutions have been designed only for a small number of categories, mostly for traffic signs associated with advanced driver-assistance systems (ADAS)~\cite{Timofte_BOOK_2011} and autonomous vehicles~\cite{Mogelmose2015phdthesis}.

Detection and recognition of a large number of traffic-sign categories remains an open question. Various previous benchmarks have addressed the traffic-sign recognition and detection task~\cite{Stallkamp2012, Houben2013,Mogelmose2012,Zaklouta2012,Zhu2016a}. However, several of them focused only on traffic-sign recognition (TSR) and ignored the much more complex problem of traffic-sign detection (TSD) where finding accurate location of traffic sign is needed. Other benchmarks that do address TSD mostly cover only a subset of traffic-sign categories, most often ones important for ADAS and autonomous vehicles applications. Most categories appearing in such benchmarks have a distinct appearance with low inter-category variance and can be detected using hand-crafted detectors and classifiers. Such examples include round mandatory signs or triangular prohibitory signs. However, many other traffic-sign classes that are not included in the existing benchmarks can be much more difficult to detect as they have a high-degree of variation in appearance. Instances of these categories may have a different real-world size, aspect ratio, color, and may contain various text and symbols (e.g., arrows) that significantly differ between instances of the same class. This often leads to a large degree of intra-category (i.e. within-category) appearance variation and at the same time leads to a low degree of inter-category (i.e. between-categories) variations due to similar appearance of objects from different categories.

Modifying existing methods with hand-crafted features and classifiers to handle such categories would be one option; however, that would be a time-consuming task, particularly when considering that many traffic-sign appearances are not consistent between countries. A much more sensible way is to use feature learning based on real examples. This can easily adapt and capture high degree of variability in appearance over a large number of traffic signs. Recent advances in deep learning have shown promising results on detection and recognition of general objects. Previous works already employed deep learning approaches for traffic-sign detection and recognition to some extent~\cite{Zhu2016}; however, their evaluation focused only on a highly limited subset of traffic-sign categories~\cite{Zhu2016a}. One of the main limitations preventing deep learning from being applied to a large set of traffic-sign categories is a lack of extensive dataset with several hundred different categories and a sufficient number of instances for each category. This issue is particularly important in deep learning where models have tens of millions of learnable parameters and large numbers of samples are needed to prevent overfitting.

In this paper, we address the issue of learning and detecting a large number of traffic-sign categories for road-based traffic-sign inventory management. As our main contribution, we propose a deep-learning-based system for training a large number of traffic-sign categories using convolutional neural networks. We base our system on the state-of-the-art detector Mask R-CNN~\cite{He2017}, which demonstrated great accuracy and speed in the field of object detection. The same network architecture is used not only for the TSR but also for accurate localization using a region proposal network, resulting in efficient end-to-end learning. In contrast to traditional approaches with hand-crafted features, the convolutional approach is applied to a broad set of categories, where individual traffic-sign instances are not only subject to change in lighting conditions, scale, viewing angle, blur, and occlusions, but also to significant intra-category appearance variations as well as low inter-category variations. Furthermore, we propose improvements to Mask R-CNN that are crucial for the domain of traffic signs. We propose adaptations that increase the recall rate, particularly for small traffic signs, and introduce a novel augmentation technique suitable for traffic-sign categories. 

As our secondary contribution, we present a novel challenging dataset with \num{200} traffic-sign categories spread over \num{13000} traffic-sign instances and \num{7000} high-resolution images. The dataset represents a novel benchmark for complex traffic signs with a large number of classes having high intra-category appearance variability. Additionally, the dataset contains enough instances to ensure appropriate learning of deep features. We achieve this by providing annotations of \num{200} traffic-sign categories with at least \num{20} instances per category (see Figure~\ref{fig:dfg-database}). Furthermore, our qualitative analysis serves as an important study for appropriateness of deep learning for the detection of large number of traffic-sign categories. 

The remainder of the paper is organized as follows. \Cref{sec:related_work} provides the related work overview, \cref{sec:method} describes the employed method, \cref{sec:experimental_results} presents the experimental results and discussion on qualitative analysis is provided in \cref{sec:qualitative_analysis}. The paper concludes with the discussion in \cref{sec:conclusion}.

\section{Related work} \label{sec:related_work}

An enormous amount of literature exists on the topics of TSR and TSD, and several review papers are available
\cite{Mogelmose2012,Wali2015_review}. In general, it is very difficult to decide which approach gives better overall results, mainly due to the lack of a standard publicly available benchmark dataset that would contain an extensive set of various traffic-sign categories, as emphasized in several recent studies \cite{Wali2015_review,Ellahyani2016}. Most authors evaluate their approaches on one of the many public datasets with a relatively limited number of traffic-sign categories:
\begin{itemize}
\item The German Traffic-Sign Detection Benchmark (GTSDB)~\cite{Houben2013}: 3 super-categories, primarily intended for detection.

\item The German Traffic-Sign Recognition Benchmark (GTSRB)~\cite{Stallkamp2012}: 43 categories, intended for recognition only.

\item The Belgium Traffic Signs (BTS) dataset~\cite{Timofte2009}: 62 categories, for detection and recognition.

\item The Mapping and Assessing the State of Traffic Infrastructure (MASTIF)~\cite{Segvic2010}: 9 original categories, extended to 31 categories \cite{Huang2016a}, acquired for road maintenance assessment service in Croatia.

\item The Swedish traffic-sign dataset (STSD)~\cite{Larsson2011}: 10 categories, for detection.

\item The Laboratory for Intelligent and Safe Automobiles (LISA) Dataset~\cite{Mogelmose2012}: 49 categories of traffic signs, acquired on the roads in the USA. 

\item The Tsinghua-Tencent 100K dataset~\cite{Zhu2016a}: 45 categories, large dataset with \num{10000} images containing at least one traffic sign and \num{90000} background images.
\end{itemize}

To enrich the set of considered traffic signs, some approaches sample images from multiple datasets to perform the evaluation \cite{Li2015,Yang2012}. On the other hand, a vast number of authors use their own private datasets \cite{LilloCastellano2015,Salti2015,Greenhalgh2012,Overett2011}. To the best of our knowledge, the largest set of categories was considered in the private dataset of \cite{Greenhalgh2012}, distinguishing between 131 categories of non-text traffic signs from the roads of United Kingdom.

Despite a large number of traffic-sign datasets, a comparison of traffic-sign detectors for large numbers of categories remains a challenging problem. In contrast to existing benchmarks that focus mostly on small numbers of super-categories (GTSDB~\cite{Houben2013}), or on small numbers of simple traffic signs (BTS~\cite{Timofte2009}, MASTIF~\cite{Segvic2010}, STSD~\cite{Larsson2011}, LISA~\cite{Mogelmose2012}), our comprehensive dataset contains 200 traffic-sign categories, including a large number of categories with significant intra-category variability. The closest large-scale dataset is the Tsinghua-Tencent 100K dataset; however, their evaluation still focuses only on 45 simple traffic signs. On the other hand, our dataset enables a comprehensive analysis of detectors in the context of traffic-sign inventory management. 

Various methods have been employed in TSR and TSD. Traditionally hand-crafted features have been used, like histogram of oriented gradients (HOG) \cite{Zaklouta2012,Greenhalgh2012,Mathias2013,Ellahyani2016,Haloi2015a,Huang2016a,Houben2013}, scale invariant feature transform (SIFT) \cite{Haloi2015a}, local binary patterns (LBP) \cite{Ellahyani2016} or integral channel features \cite{Mathias2013}. A wide range of machine learning methods have also been employed, ranging from support vector machine (SVM)~\cite{Greenhalgh2012,Ellahyani2016,Zaklouta2014}, logistic regression \cite{Pei2013}, and random forests \cite{Ellahyani2016,Zaklouta2014}, to
artificial neural networks in the form of an extreme learning machine (ELM) \cite{Huang2016a}.

Recently, like the entire computer vision field, TSR and TSD has also been subject to CNN renaissance. A modern CNN approach that automatically extracts multi-scale features for TSD has been applied in~\cite{Wu2013}. In TSR, CNNs have been used to automatically learn feature representations as well as to perform the final classification~\cite{Ciresan2012,Sermanet2011,Jin2014,Vukotic2014}.
In order to further improve the recognition accuracy, a combination of CNN and Multilayer Perceptron was applied in~\cite{Ciresan2011}, while an ensemble classifier consisting of several CNNs was proposed in~\cite{Ciresan2012,Jin2014}. A method that uses CNN to learn features and then applies ELM as a classifier has been applied in~\cite{Zeng2015}, while \cite{Haloi2015} employed a deep network consisting of spatial transformer layers and a modified version of inception module. It has been shown in~\cite{Stallkamp2011} that the performance of CNN on recognition outperforms the human performance on GTSRB. A combined problems of TSR and TSD were addressed using CNNs in recent works of~\cite{Zhu2016,Zhu2016a}. In the latter, they use a heavily modified OverFeat~\cite{Sermanet} network, while in the former they applied a fully convolutional network to obtain a heat map of the image, on which a region proposal algorithm was employed for detection. Finally, a separate CNN was then employed to classify the obtained regions.

Our proposed deep-learning-based approach differs from previous related works. In contrast to traditional approaches with hand-crafted features and machine learning~\cite{Zaklouta2012,Greenhalgh2012}, we propose full feature learning with end-to-end learning. Our approach also differs from other deep-learning-based traffic-sign detection methods. Our method, which is based on Mask R-CNN, uses region proposal network instead of using a separate method for generating region proposals as in~\cite{Zhu2016}, and in contrast to~\cite{Zhu2016a}, we employ deeper networks based on the VGG16~\cite{Simonyan2015} and ResNet-50~\cite{He2015a} architectures. As opposed to both~\cite{Zhu2016} and~\cite{Zhu2016a}, we also employ network pre-trained on ImageNet, which significantly reduces the need for training samples. In addition, we have implemented several extensions leading to superior performance.

\section{Traffic-sign detection with Mask R-CNN} \label{sec:method}

In this section, we present our system for traffic-sign detection using the Mask R-CNN detector extended with several improvements. First, we present the original Mask R-CNN detector, then we present our adaptation for learning traffic-sign categories, and finally, we present our data augmentation technique.

\subsection{Mask R-CNN}\label{sec:maskRcnn}
Here we briefly describe Mask R-CNN and refer the reader to~\cite{He2017} for a more detailed description. The Mask R-CNN network~\cite{He2017} is an extension of Faster R-CNN~\cite{fasterRCNN}, both of which are composed of two modules. The first module is deep fully convolutional network, a so-called Region Proposal Network (RPN), that takes an input image and produces a set of rectangular object proposals, each with an objectness score. The second module is a region-based CNN, called Fast R-CNN, that classifies the proposed regions into a set of predefined categories. Fast R-CNN is highly efficient, since it shares convolutions across individual proposals. It also performs bounding box regression to further refine the quality of the proposed regions. The entire system is a single unified network, in which RPN and Fast R-CNN are merged by sharing their convolutional features. Following the recently popular terminology of neural networks with the \textit{``attention''} mechanism, the RPN module tells the Fast R-CNN module where to look. Mask R-CNN then improves this system by  combining the underlying network architecture with a Feature Pyramid Network (FPN)~\cite{Lin2016b}. With the FPN, the detector is able to improve the performance on small objects, since FPN extracts features from lower layers of the network, before the down-sampling removes important details in small objects. The underlaying network architecture, which is VGG16~\cite{Simonyan2015} in Faster R-CNN, is replaced with a residual network (ResNet)~\cite{He2015a} in Mask R-CNN.

Faster and Mask R-CNN are trained for the region proposal task as well as for the classification task. This is performed with a stochastic gradient descent. Mask R-CNN learns both networks simultaneously using end-to-end learning. The original Faster R-CNN implementation performed this with a 4-step optimization process that alternated between the two tasks. However, the newer end-to-end learning scheme from Mask R-CNN is also applicable to Faster R-CNN. Commonly, both networks are initialized with the ImageNet pre-trained model before they are trained on the specific domain.

Both methods enable fast detection and recognition in the test-phase. For each input image the trained model outputs a set of object bounding boxes, where each box is associated with a category label and a softmax score in the interval of $[0, 1]$.
 
\subsection{Adaptation to traffic-sign detection}


Mask R-CNN is a general method developed for the detection and recognition of general objects. In order to adapt it to the particular domain of TSD, we developed and implemented several domain specific improvements.

\paragraph{Online hard-example mining} We first incorporate online hard-example mining (OHEM) into the classification learning module (Fast R-CNN module). Following the work of Shrivastava et al.~\cite{Shrivastava2016}, that introduced OHEM for Faster R-CNN, we replace the method for selecting regions of interest (ROIs) that are passed to the classification learning module. Normally, \num{256} ROIs per image are selected randomly, some as foreground (traffic signs) and some as background (non-traffic signs). In our approach, we replace random selection of ROIs with the selection based on their classification loss value. Regions are sorted based on their loss value and only ones with high enough loss are passed to the classification learning module. This ensures learning on samples on which the network was mistaken the most, i.e., on hard examples. We perform selection separately for the background and the foreground objects to ensure sufficient positive and negative samples during each gradient descent step. 

We implement OHEM as an end-to-end learning by utilizing the existing classification module to obtain the classification losses for ROIs. Note that classification loss, which represents a criteria for selecting ROIs, is not computed for all possible ROIs generated by the RPN but only for the top ROIs based on their objectness score. We take \num{2000} regions and perform a non-maxima suppression (NMS) to eliminate duplicated ROIs. This is a standard approach to reduce the number of ROIs in Mask R-CNN before they are selected for learning. We experimented with using more than \num{2000} regions before the NMS but this significantly increased the learning time due to slower NMS without contributing to any performance gain.

\paragraph{Distribution of selected training samples} The mechanism for selecting the training samples for the region proposal network is also improved in the proposed approach. Originally, the Mask R-CNN selects ROIs randomly. This is done separately for foreground and background. However, when many small and large objects are present in the image at the same time the random selection introduces imbalance into the learning process. The imbalance arises due to large objects having a large number of ROIs that cover it, while small objects having only a small number of ROIs. Selecting samples based on this distribution will skew the learning process, since larger objects will be observed more often and favored much more than the smaller ones. To alleviate this issue we change the distribution of the selected training samples to evenly cover all sizes of the training objects. We achieve this by selecting the same number of ROIs for each object present in the image.

\paragraph{Sample weighting} We incorporate additional weighting of samples during the learning process. Our evaluation showed that Mask R-CNN cannot achieve 100\% recall due to missing region proposals in certain cases. We address this issue with different weighting of the training regions. During the learning, both foreground and background regions are selected; however, there are often many more background regions, since most traffic signs in images are small and only a few region proposals exists for those traffic signs. Without any weighting the learning process will observe background objects more often and will focus on learning the background instead of on the foreground. We address this problem with smaller weights for the background regions, which forces the network to learn foreground objects first. This is implemented for the training process of the region proposal network as well as for the classification network, weighting backgrounds with \num{0.01} for the RPN and \num{0.1} for the classification network. This improvement is particularly important for the RPN, since regions missed at this point in the pipeline cannot be recovered later by the classification module and would lead to poor overall recall if not addressed.

\paragraph{Adjusting region pass-through during detection} Lastly, we also change the number of ROIs passed from the RPN to the classification network during the detection stage. The number of regions passed through need to be adjusted due to a large number of small objects that are commonly present in the traffic-sign domain. We increase this number from \num{1000} to \num{10000} regions per one FPN level before the NMS. After merging ROIs from all FPN levels and performing the NMS \num{2000} regions are retained.

\subsection{Data augmentation} \label{sec:aug}

An important factor to consider when learning deep models is the size of the training set. Due to millions of learnable parameters the system becomes undetermined without a sufficient number of training samples. We partially address this issue with a pre-trained model, one learned on \num{1.2} million images of ImageNet, but we also propose an additional data augmentation. The nature of the traffic-sign domain allows us to construct a large number of new samples using artificial distortions of existing traffic-sign instances.

An additional synthetic traffic-sign instances are created by modifying segmented, real-world training samples. The traffic signs in the proposed dataset are annotated with tight bounding boxes (see Figure~\ref{fig:dfg-annot}), allowing to be segmented from the training images. Two classes of distortions were performed: (i) geometric/shape distortions (perspective change, changes in scale), and (ii) appearance distortions (variations in brightness and contrast).

Before applying geometric and appearance distortions we first normalized each traffic-sign instance. For the appearance normalization, we normalized contrast of the intensity channel in the L*a*b domain, while for the geometric normalization, we calculated the homography between instance annotation points and a geometric template for a specific traffic-sign class. We manually created templates for most of the classes with the exception of several classes where this was not possible (e.g. the train crossing sign, direction signs with the shape of an arrow, etc.). We generated new synthetic instances for those classes as well but without performing geometry normalization and without applying geometric distortions to synthetic instances.

\begin{figure}
\includegraphics[width=\columnwidth]{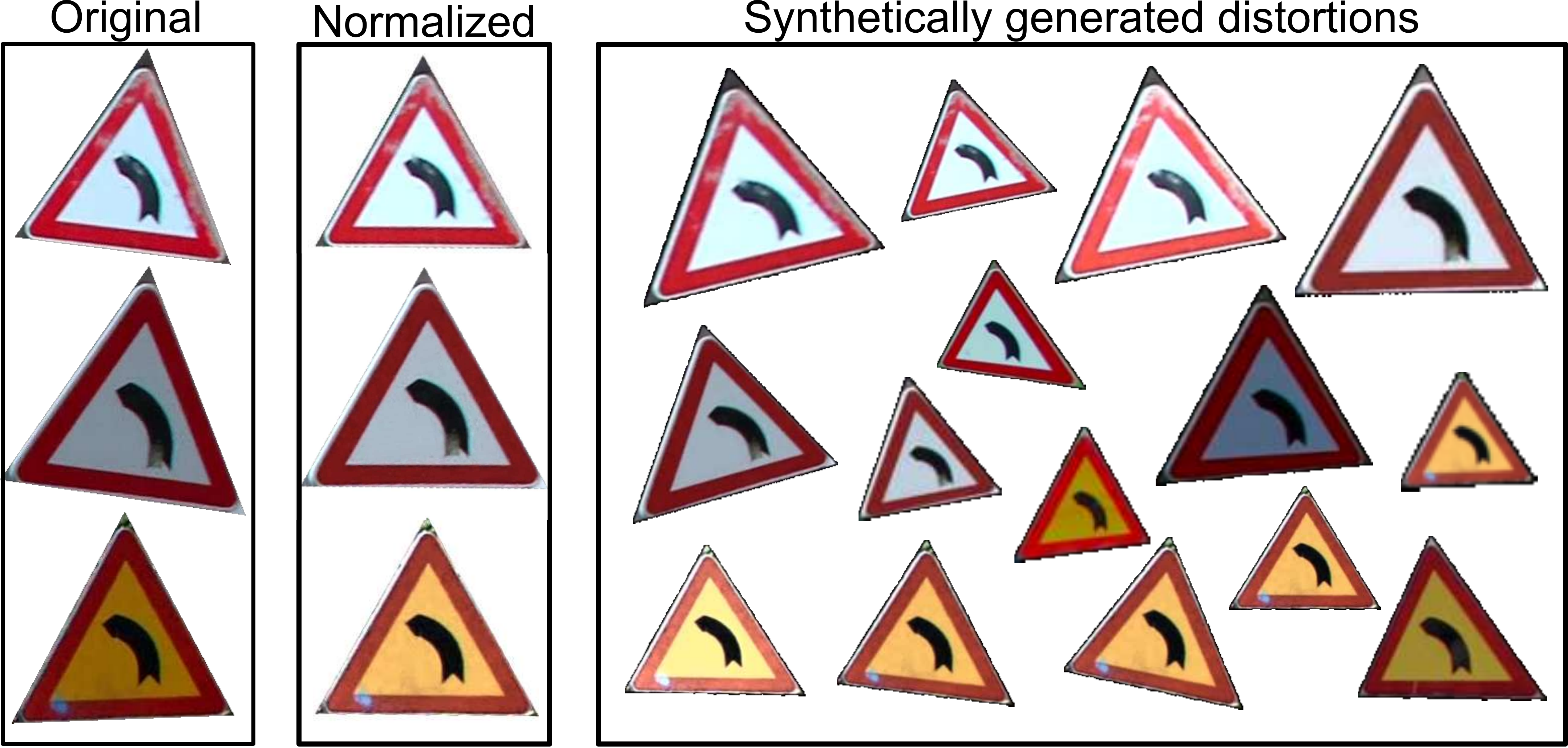}
\caption{Several examples of traffic-sign instances as generated during the process of data augmentation: (a) original image on the left, (b) normalized geometry and appearance in the middle, and (c) generated samples with synthetic distortions on the right.\label{fig:aug-overview} }
\end{figure}

\begin{figure}
\includegraphics[width=\columnwidth]{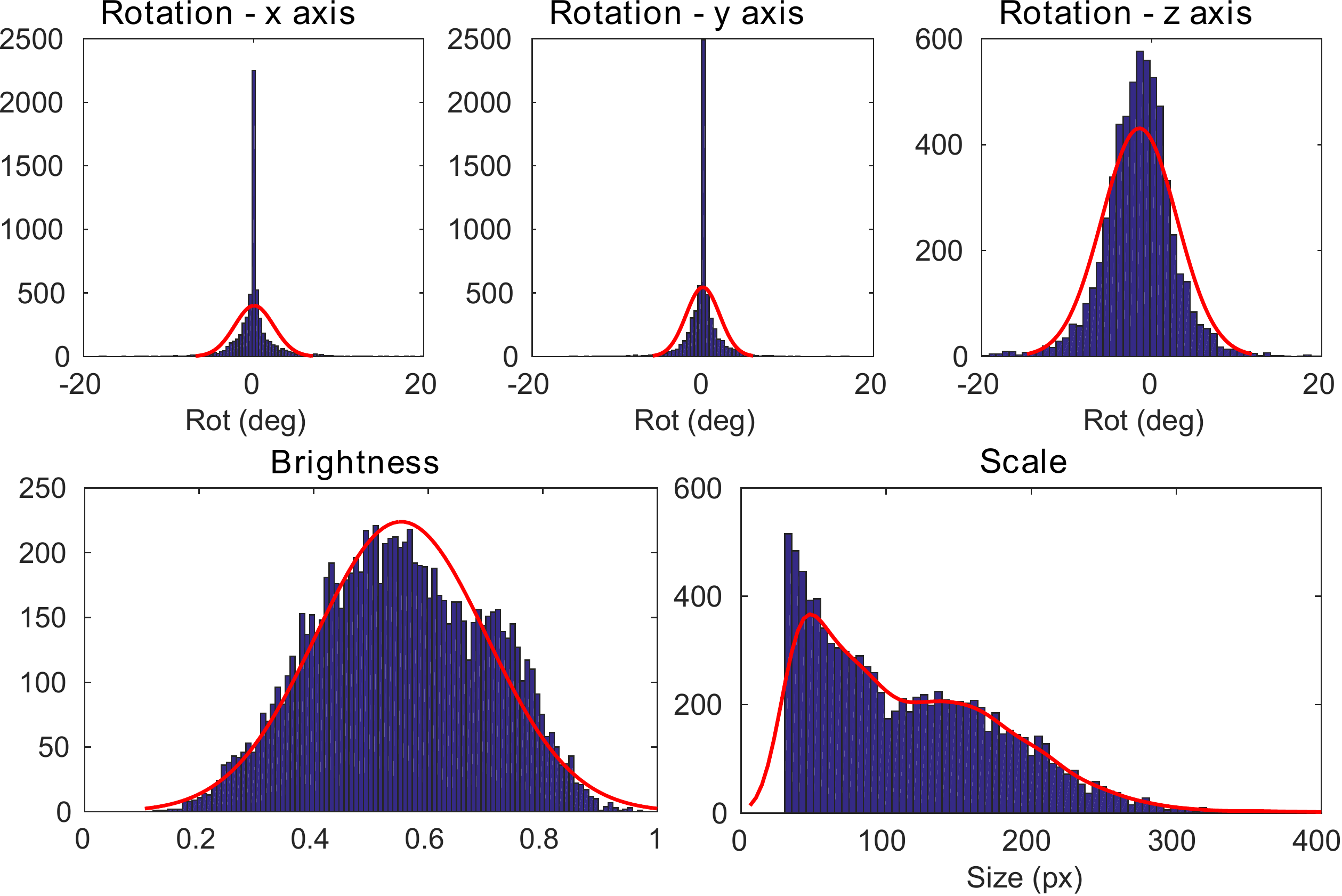}
\caption{Distributions of traffic-sign distortions computed for rotation in the top row, appearance (i.e. brightness) in the bottom left side and scale in the bottom right side. Red lines represent the Gaussian distributions, which are sampled when  generating new examples. \label{fig:aug-distr}}
\end{figure}

In order to generate synthetic training samples that are as realistic as possible, we followed the distribution of the training set's geometry and appearance variability. For the geometry change we estimated the distribution of Euler rotation angles (in X,Y and Z axis) of trainings examples, while for the appearance change, we estimated the distribution of averaged intensity values. We additionally estimated the distribution of scales using the size of geometry normalized (rectified) instances. We modeled all changes with a Gaussian mixture model, but used a single mixture component, K=1, for the geometry and appearance, and two mixture components, K=2, for the scale. Several examples of original, normalized and synthetically generated samples are shown in Figure~\ref{fig:aug-overview}, while a histogram and its corresponding distributions for different distortions are depicted in Figure~\ref{fig:aug-distr}.

When generating synthetic distortions we sampled random values from the corresponding distributions. However, variance that is twice as large as the variance in the observed distribution was used to increase the likelihood of generating larger distortions. In the appearance distortion the distributions were not generic for all classes, but instead, we used different distribution for each classes. We used class specific mean instead of mean over all categories but we still applied common variance calculated from all the categories. This guarded us from generating invalid contrast values for very dark/bright categories, such as gray or white direction signs.

To emulate the real-world settings, the newly generated traffic-sign instances were inserted into the street-environment-like background images. Background images were acquired from the subset of the BTS dataset \cite{Timofte2009}, which contains no other traffic signs. At least two, and at most five, traffic signs were placed in a non-overlapping manner in random locations of each background image, avoiding the bottom central part where only the road is usually seen. With the whole augmentation process we generated enough new instances to ensured each category has at least \num{200} instances. This resulted in around \num{30000} new traffic-sign instances spread over \num{8775} new training images. 

\section{The DFG traffic-sign dataset} \label{sec:dataset}

Our dataset was acquired by the DFG Consulting d.o.o. company for the purpose of maintaining inventory of traffic signs on Slovenian roads. The RGB images were acquired with a camera mounted on a vehicle that was driven through several different Slovenian municipalities. The image data was acquired in rural as well as in urban areas. Only images containing at least one traffic sign were selected from the vast corpus of collected data. Moreover, the selection was performed in such a way that there is usually a significant scene change between any pair of selected consecutive images. Since images were acquired for the purpose of maintaining traffic-sign inventory, this allowed the image acquisition to be performed in the day-time avoiding bad weather conditions such as rain, snow and fog. Nevertheless, the dataset does include other difficult variations in the weather and the environment that are present in the real-world environment such as: rural and city/urban landscape, different levels of natural occlusions and shadows, and various ranges of a cloudy sky and direct sunlight. Images taken under winter conditions with snow cover were also included.

\begin{figure}
\centering
\includegraphics[width=0.9\columnwidth]{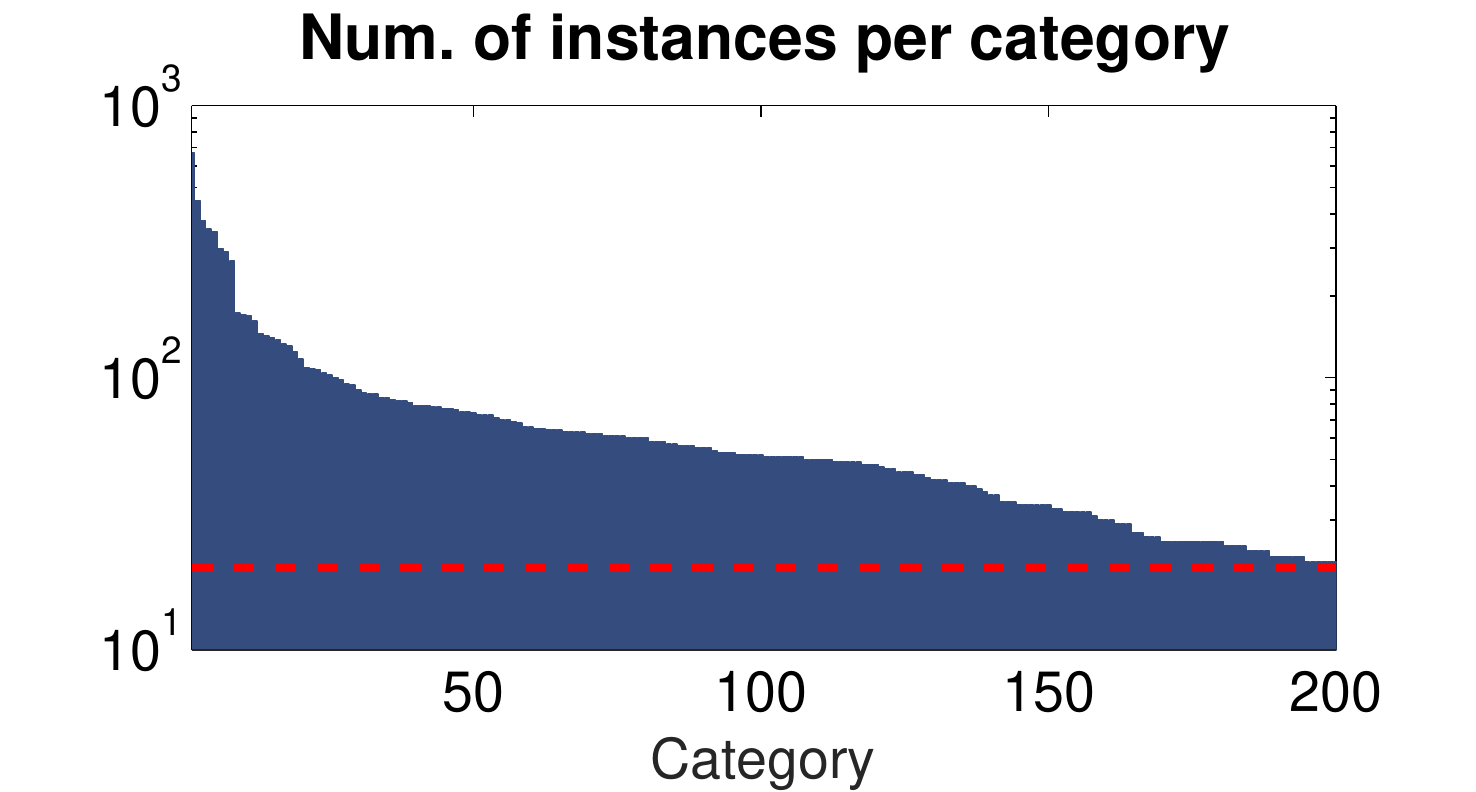}
\caption{Distribution of number of instances over categories in the DFG traffic-sign dataset. Horizontal red dashed line represents 20 instances per category, which we use as a cut-off point. Note, the distribution is shown in the logarithmic scale. \label{fig:dfg-distribution} }
\end{figure}

\begin{figure}
\centering
\includegraphics[width=0.85\columnwidth]{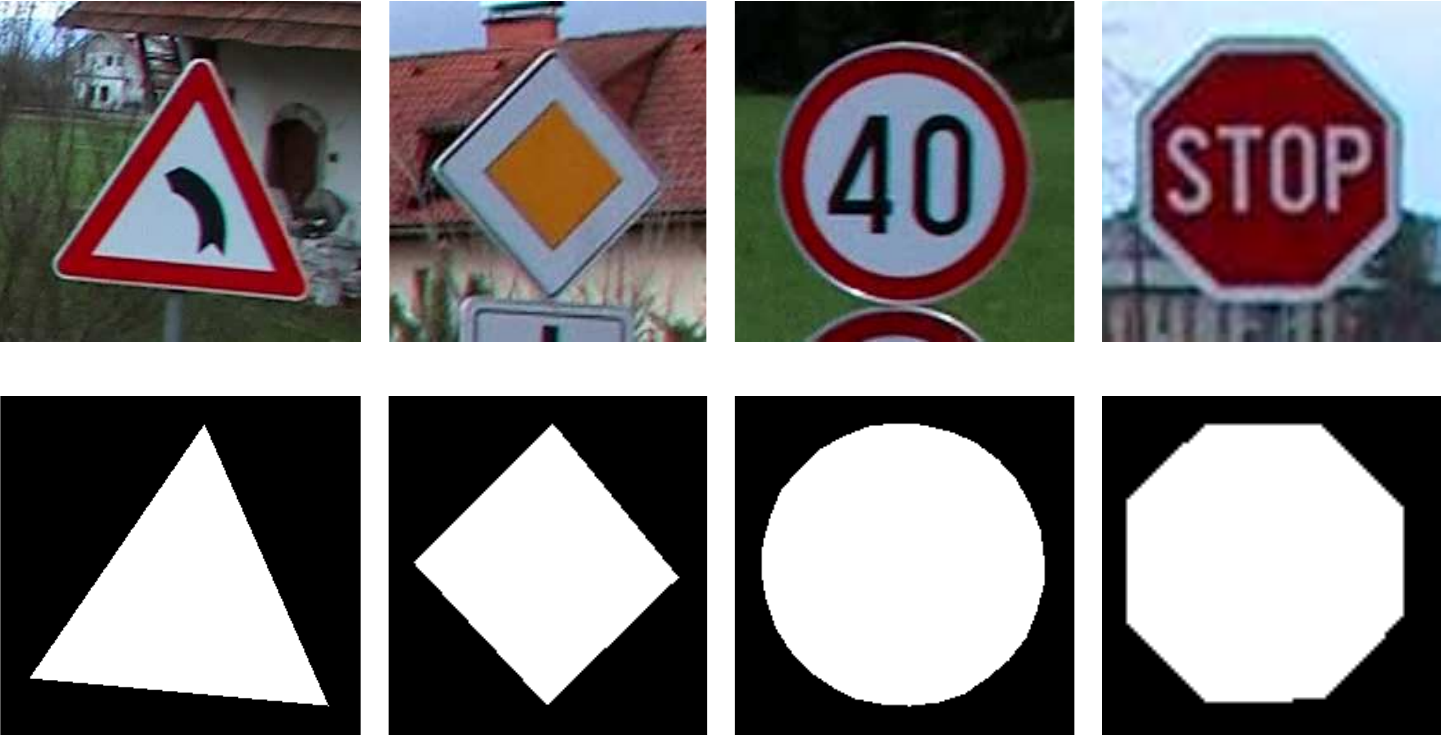}
\caption{Several examples of traffic signs in the DFG traffic-sign dataset with their corresponding annotation masks showing the precision of the annotation mask. \label{fig:dfg-annot} }
\end{figure}

The dataset, termed the DFG traffic-sign dataset\footnote{The dataset, termed DFG traffic-sign dataset, is publicly available at http://www.vicos.si/Downloads/DFGTSD}, contains a total of \num{6957} images with \num{13239} tightly annotated traffic-sign instances corresponding to \num{200} categories. The total number of instances is different for each category (see Figure~\ref{fig:dfg-distribution}). Each image contains annotations of all traffic signs larger than 25 pixels for any of the \num{200} categories in a tightly annotated polygon (see Figure~\ref{fig:dfg-annot}). Categories in the dataset represent a subset of all categories from the corpus of raw images provided by the company; however, some categories in the corpus did not meet the necessary criteria to create a quality dataset. In particular, all categories in the public dataset now meet the following three criteria: (a) each category has a sufficient number of instances (at least 20 instances with a minimal bounding box size of 30 pixels), (b) each category represents a planar object and (c) each category contains traffic signs that have at least some visual consistency. Among all categories in the DFG traffic-sign dataset roughly \SI{70}{\%} of them correspond to traffic signs with low appearance changes, while a significantly larger appearance variability is present in the remaining \SI{30}{\%}. Latter signs can be of variable aspect ratio or color and can contain various text and numbers. See 200 categories of traffic signs depicted in Figure~\ref{fig:dfg-database}.

Note that the dataset contains annotations as small as 25 pixels. However, annotations smaller than 30 pixels are flagged as difficult and are not considered neither for the training nor for the testing. We selected 30 pixels as a minimal size based on down-sampling of features in Faster and Mask R-CNN, which is performed 5-times and results in 32x32 pixels being represented by 1x1 feature pixel.

A suitable train-test split was generated to provide a sufficient number of samples for both the training and the test set. A restriction was set that \SI{25}{\%} of traffic-sign instances for each category have to appear in the test set. For the smallest categories with only \num{20} instances, this ensured a minimum number of 15 samples for the training set and a minimum number of 5 samples for the test set. Images were assigned randomly to either the training or the test set. However, additional constraint mechanism was employed to ensure all images of the same physical object are always present either in the test set or in the training set but never in both of them at the same time. This was ensured by clustering images within 50 meter distance and assigning whole clusters to the training or the test set. In this way, we generated a training set with \num{5254} images and a test set with \num{1703} images.

\section{Experimental evaluation} \label{sec:experimental_results}

In this section, we perform extensive evaluation of deep learning methods that are appropriate for the traffic-sign detection and recognition. We focus on evaluating two state-of-the-art, region-proposal-based methods: Faster R-CNN and Mask R-CNN. We first perform evaluation on the existing public traffic-sign dataset to establish a baseline comparison with the related work. Swedish traffic-sign dataset (STSD) is used for this purpose. Then, an extensive evaluation on newly proposed DFG traffic-sign dataset is performed with a comprehensive analysis of the proposed improvements.

\subsection{Implementation details} \label{sec:implementation}

\begin{table*}[!b]
\centering

\caption{Detailed results on Swedish traffic-sign dataset (STDS) for different categories.\label{tab:STSD-details}}
\small{
\begin{tabular}{l rrp{5pt} rrrp{0pt}p{-5pt} rrrp{1pt}p{1pt} rrr}
\toprule
\multirow{6}{*}{\textit{Traffic Sign}} & \multicolumn{2}{l}{FCN~\cite{Zhu2016}} && \multicolumn{3}{l}{Faster R-CNN} &&& \multicolumn{7}{l}{Mask R-CNN (ResNet-50)} \vspace{2pt}\\
\cline{2-3} \cline{5-7} \cline{10-17} \\
 & &&&&&&&& \multicolumn{3}{c}{No adaptations} &&& \multicolumn{3}{c}{With adaptations (our)} \vspace{3pt} \\
 & Prec.& Rec. && Prec.& Rec.& AP$^{50}$ &&& Prec.& Rec. & AP$^{50}$ &&& Prec.& Rec.& AP$^{50}$ \\
 \midrule
\makecell[l]{PED. CROS.} 		& 100.0 & 95.2 	&& 92.6 & 92.6 & 94.1 & 	&& 100.0 & 97.5 & 98.2 & 	&& 99.2 & 97.6 & 97.6 \\
PASS RIGHT SIDE				& 95.3 & 93.8 	&& 98.1 & 98.1 & 99.5 & 		&& 94.8 & 98.2 & 98.6 & 		&& 100.0 & 98.2 & 99.8 \\
\makecell[l]{NO STOP/STAN} 	& 100.0 & 75.0 	&& 92.3 & 92.3 & 86.5 & 		&& 81.2 & 100.0 & 95.4 & 	&& 86.7 & 100.0 & 83.9 \\
50 SIGN 					& 100.0 & 100.0 && 81.2 & 92.9 & 90.3 & 	&& 87.5 & 100.0 & 97.5 & 	&& 90.0 & 96.4 & 96.9 \\
PRIORITY ROAD 				& 100.0 & 98.9 	&& 98.7 & 95.1 & 92.1 & 		&& 97.5 & 97.5 & 96.9 & 		&& 98.7 & 92.9 & 89.8 \\
GIVE WAY 					& 96.7 & 96.7 	&& 100.0 & 94.1 & 94.1 & 	&& 100.0 & 91.4 & 91.4 & 	&& 100.0 & 94.1 & 94.1 \\
70 SIGN 					& 100.0 & 100.0 && 100.0 & 100.0 & 100.0 & 	&& 100.0 & 100.0 & 100.0 & 	&& 100.0 & 100.0 & 100.0 \\
80 SIGN 					& 94.4 & 77.3 	&& 100.0 & 95.2 & 95.2 & 	&& 95.2 & 100.0 & 99.8 & 	&& 100.0 & 100.0 & 100.0 \\
100 SIGN 					& 90.5 & 100.0	&& 94.1 & 88.9 & 92.5 & 		&& 100.0 & 61.1 & 74.8 & 	&& 100.0 & 93.8 & 93.8 \\
NO PARKING 					& 100.0 & 92.1 	&& 96.8 & 90.9 & 98.5 & 		&& 96.7 & 90.6 & 95.9 & 		&& 100.0 & 93.9 & 96.5 \\
\vspace{-5pt}&&&&&&&&&&&\\
\textit{Averaged}  	& \textbf{97.7} & 92.9 &&	95.4 & 94.0 & 94.3 & 		&& 95.3 & 93.6 & 94.9 & 		&& 97.5 & \textbf{96.7} & \textbf{95.2} \\
\bottomrule
\end{tabular}
}
\end{table*}

A publicly available Caffe2-based, Python implementation of the Detectron~\cite{Girshick2018} is used for both Faster and Mask R-CNN\footnote{Our proposed improvements have been implemented in the Detectron framework and are publicly available in the GitHub repository: https://github.com/skokec/detectron-traffic-signs}. For the Faster R-CNN, we employ the VGG16~\cite{Simonyan2015} network with 13 convolutional layers and 3 fully-connected layers, while for the Mask R-CNN, we employ a residual network~\cite{He2015a} with 50 convolutional layers (ResNet-50). The ResNet-50 architecture consists of 16 convolutional filters with kernel sizes of $3\times 3$ or larger. Mask R-CNN also implements Feature Pyramid Network (FPN)~\cite{Lin2016b}, which collects features from different layers of the network to capture the information from small objects, which may be removed in higher layers due to down-sampling. Both networks are initialized with a model pre-trained on ImageNet as provided by~\cite{Girshick2018}. We also experimented with larger variant of the residual network using 101 layers (ResNet-101), but performance did not improve compared to ResNet-50. We therefore focused only on the ResNet-50, which at the same time is faster with half the layers of ResNet-101.

Both methods use similar learning hyper-parameters. A learning rate of \num{0.001} is used for Faster R-CNN with a weight decay of \num{0.0005}, while a learning rate of \num{0.0025} and a weight decay of \num{0.0001} is used for Mask R-CNN. Both approaches also use momentum of \num{0.9}. The same hyper-parameters are used in all experiments. Note that the same hyper-parameters are used in~\cite{Girshick2018} to pre-train the model on ImageNet dataset. Both methods are trained end-to-end with simultaneous learning of both the region proposal network and the classification network. We learn both methods for 95 epochs and reduce the learning rate by a factor of 10 at the 50th and 75th epoch. We use two images per batch per GPU and train on STSD with 2 GPUs and on DFG dataset with 4 GPUs. This resulted in effectively using 4 images per batch on the STSD and 8 images per batch on the DFG dataset. 

\subsection{Performance metrics}

Several different metrics are used in this study to evaluate the proposed approach. As a primary metric, we report mean average precision (mAP), which is commonly used in the evaluation of visual object detectors. We use two variants of the mAP: (i) mAP$^{50}$, based on the PASCAL visual object challenge~\cite{pascal-voc-2008}, and (ii) mAP$^{50:95}$, based on the COCO challenge~\cite{Lin2014}. Both metrics define a minimal intersection-over-union (IoU) overlap with the groundtruth region for a detection to be considered as a true positive, and both compute average precision (AP) as the area under the precision-recall curve to accurately capture the trade-off between the miss rate and the false-positive rate. AP is calculated for each category independently and the final metric consists of AP values averaged over all categories. A fixed IoU overlap is used in the mAP$^{50}$---using the PASCAL-based IoU overlap of 0.50---however, in mAP$^{50:95}$, the reported value is an average of mAP values calculated at a range of IoU overlap values. The reported values are averaged over the IoU overlap range of $[0.50,0.95]$ with \num{0.05} increments, the same range as used in the COCO detection challenge~\cite{Lin2014}. Thus, the COCO-based mAP gives more emphasis on the quality of region overlaps, while the PASCAL-based mAP ignores that aspect.

\begin{table}
\centering
\caption{Evaluation on Swedish traffic-sign dataset (STSD) with reported averaged values over ten categories.\label{tab:STSD-avg}}
\small{
\begin{tabular}{p{39pt} ccccccc}
\toprule
\multirow{3}{*}{\textit{Average}} & \multirow{3}{*}{\makecell[c]{R-CNN\\\cite{Zhu2016}} } &  \multirow{3}{*}{\makecell[c]{FCN\\\cite{Zhu2016}} }  &  \multirow{3}{*}{\makecell[l]{Faster\\R-CNN}} & \multicolumn{2}{l}{\makecell[l]{Mask R-CNN\\(ResNet-50)}} \\
 \cline{5-6} 
\vspace{-7pt}\\
 & & & & No adapt. & \makecell[l]{Adapt.\\(ours)}  \\ 

\midrule
Precision	& 91.2	& \textbf{97.7}	& 95.4	& 95.3	 & 97.5\\
Recall		& 87.2	& 92.9	& 94.0	& 93.6	& \textbf{96.7}\\
F-measure   & 88.8	& 95.0	& 94.6	& 93.8	& \textbf{97.0}\\
mAP$^{50}$  &  /	& /		& 94.3	& 94.9	& \textbf{95.2}\\
\bottomrule
\end{tabular}
}

\end{table}


For comparison with the state-of-the-art, we also report precision and recall values at best F-measure and their corresponding error rates, i.e.  false-positive rate as $1-precision$ and miss rate as $1-recall$, respectively. The false-positive rate shows how many detections are false, while the miss rate reveals how many traffic signs were not detected at all.

\begin{figure}
\centering
\vspace{-5pt}
\includegraphics[width=0.9\columnwidth]{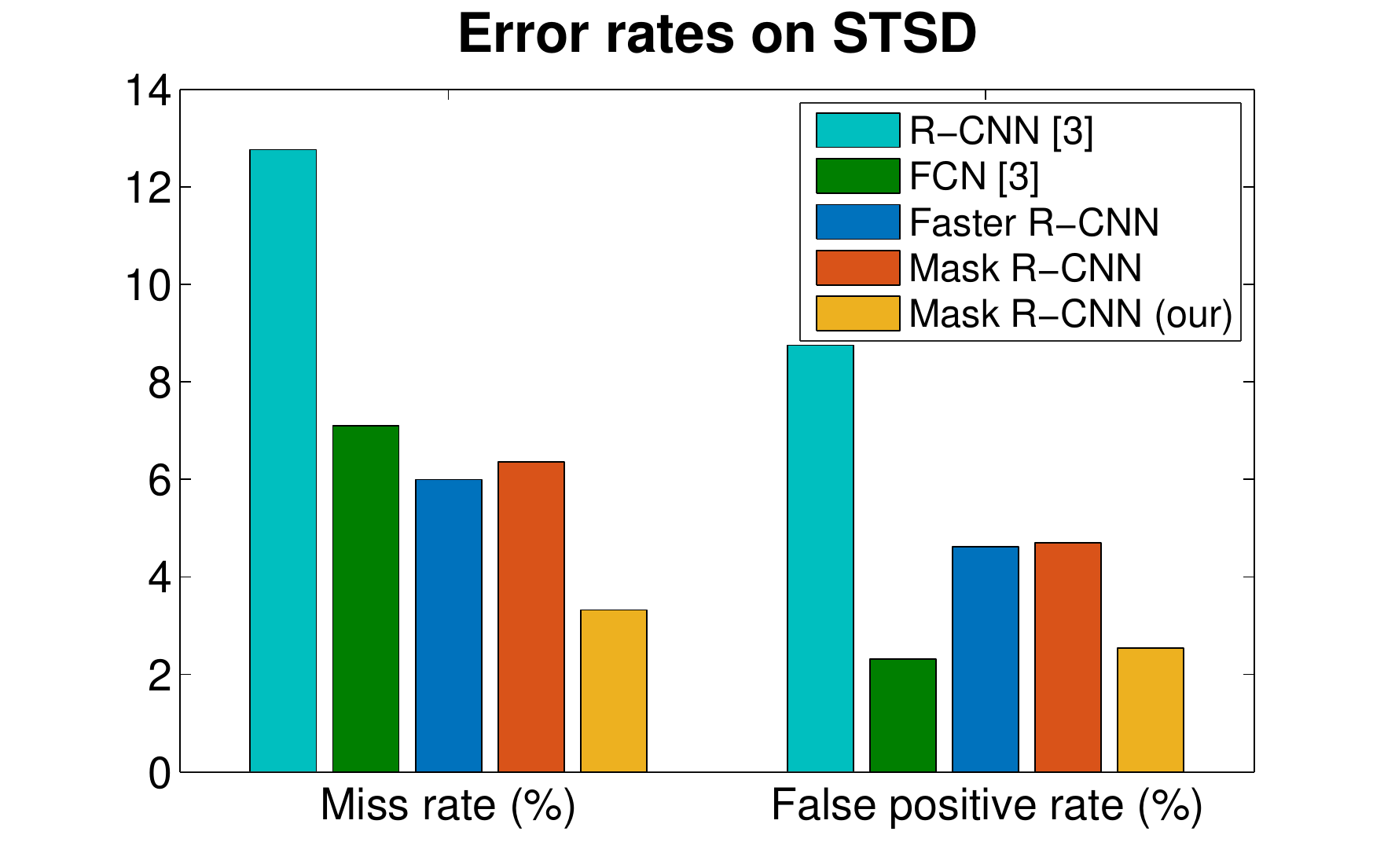}
\caption{Miss rates ($1-recall$) and false positive ($1-precision$) rates on Swedish traffic-sign dataset averaged over ten categories. Values are calculated at ideal F-measure. Note, smaller values are better.}\label{fig:STSD-graph}
\end{figure}
\subsection{Comparison to the state-of-the-art}

Although many previously proposed approaches exist, it is quite difficult to perform a reliable comparison with those approaches, since they are mostly evaluated on non-public datasets or, only on the TSR task. To this end, we evaluated the proposed method on the Swedish traffic-sign dataset (STSD), comparing the results to the previously best performing methods published in~\cite{Zhu2016}, and indirectly to other methods reported therein.

\begin{figure*}
\centering
\subfloat[]{\includegraphics[width=0.25\textwidth]{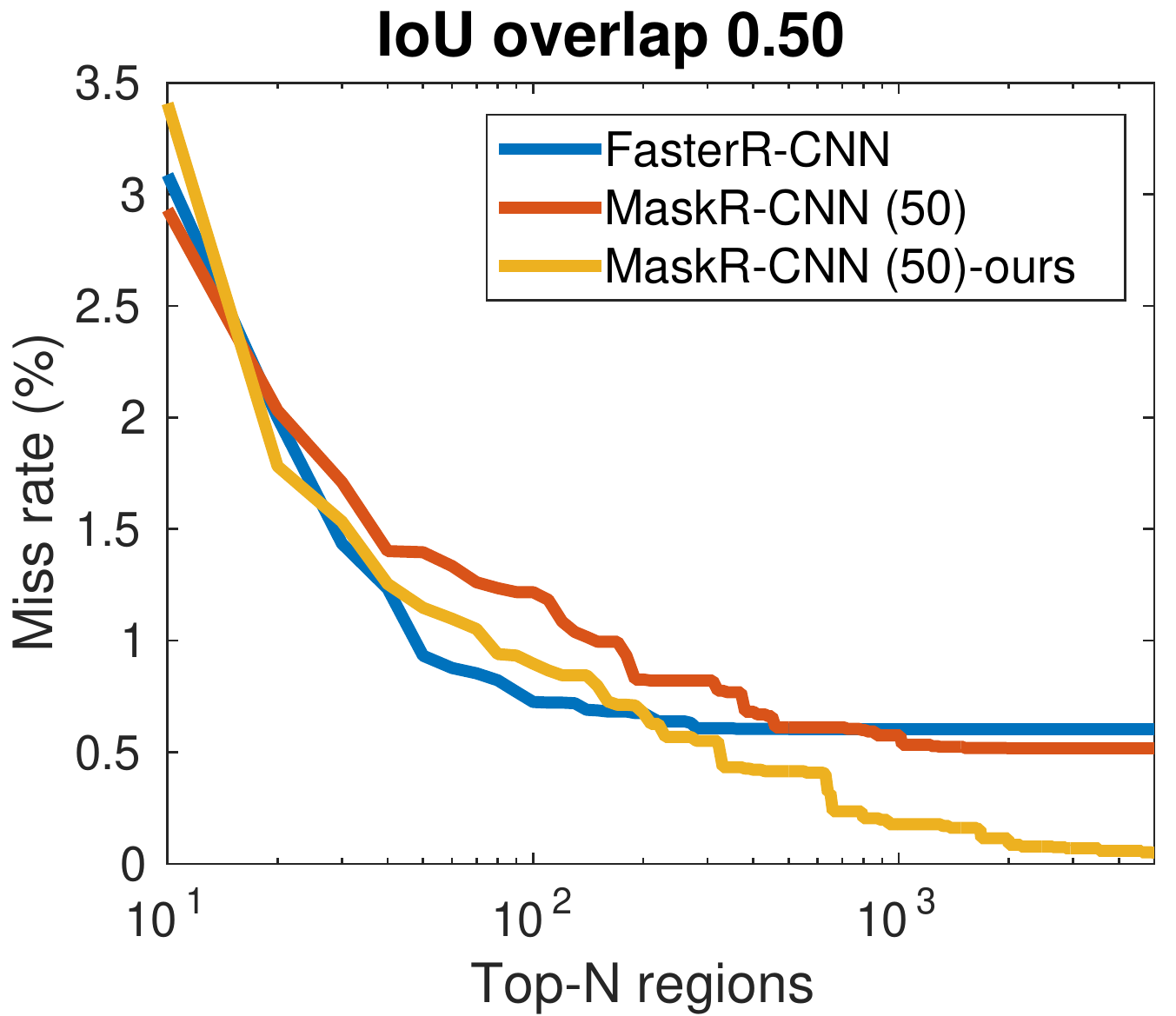}\label{fig:rpn-a}}
\subfloat[]{\includegraphics[width=0.25\textwidth]{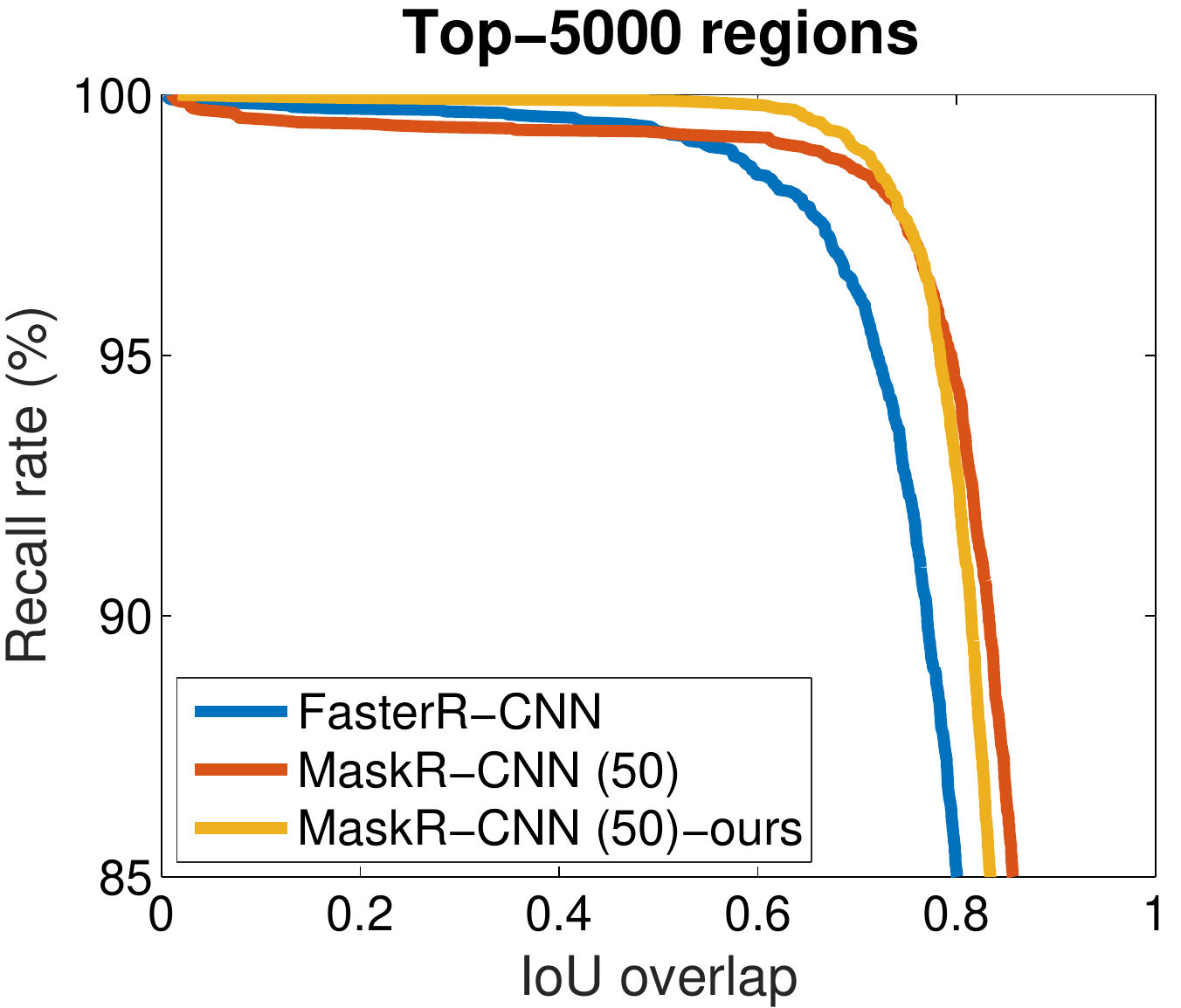}\label{fig:rpn-b}}
\subfloat[]{\includegraphics[width=0.25\textwidth]{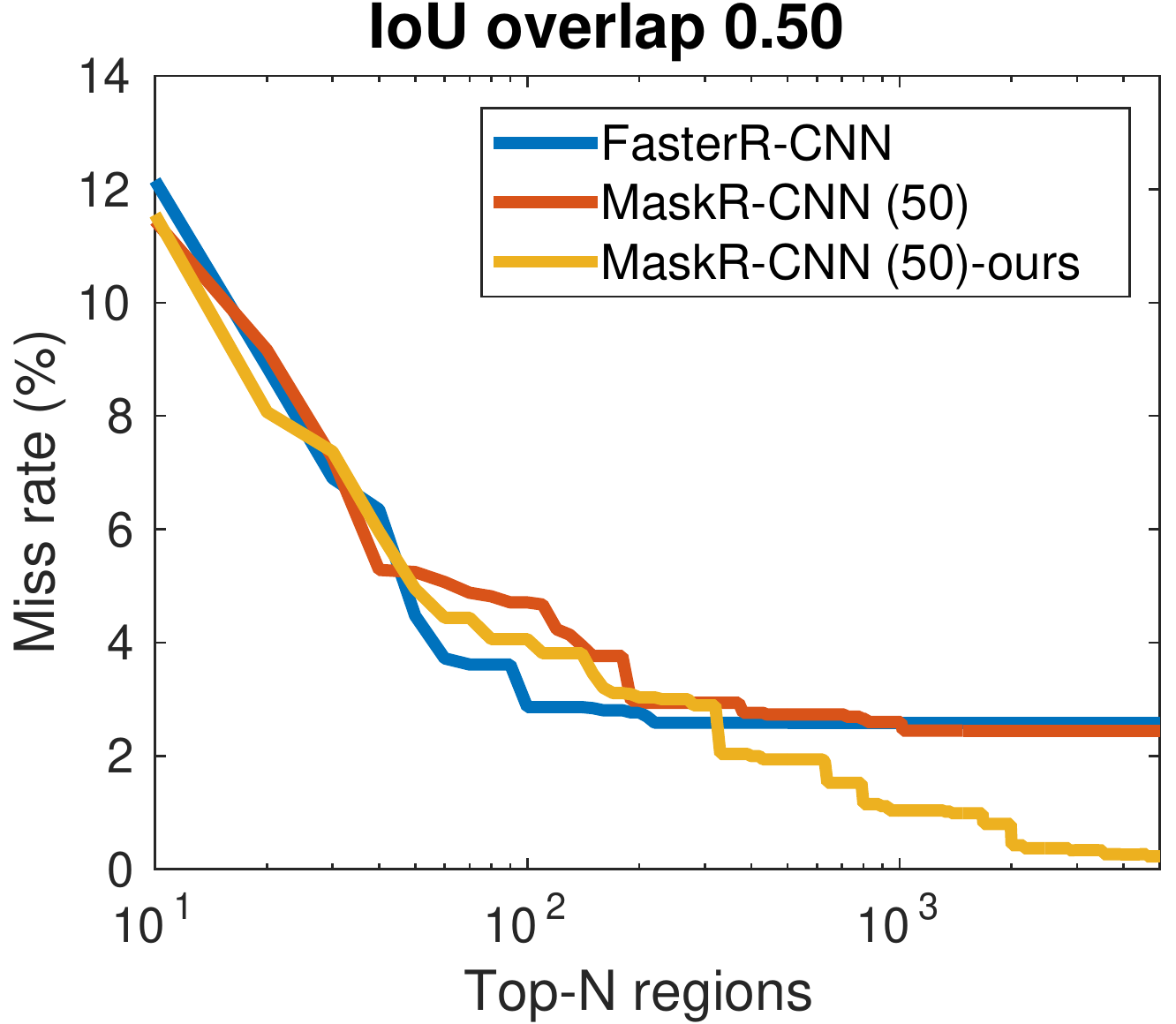}\label{fig:rpn-c}}
\subfloat[]{\includegraphics[width=0.25\textwidth]{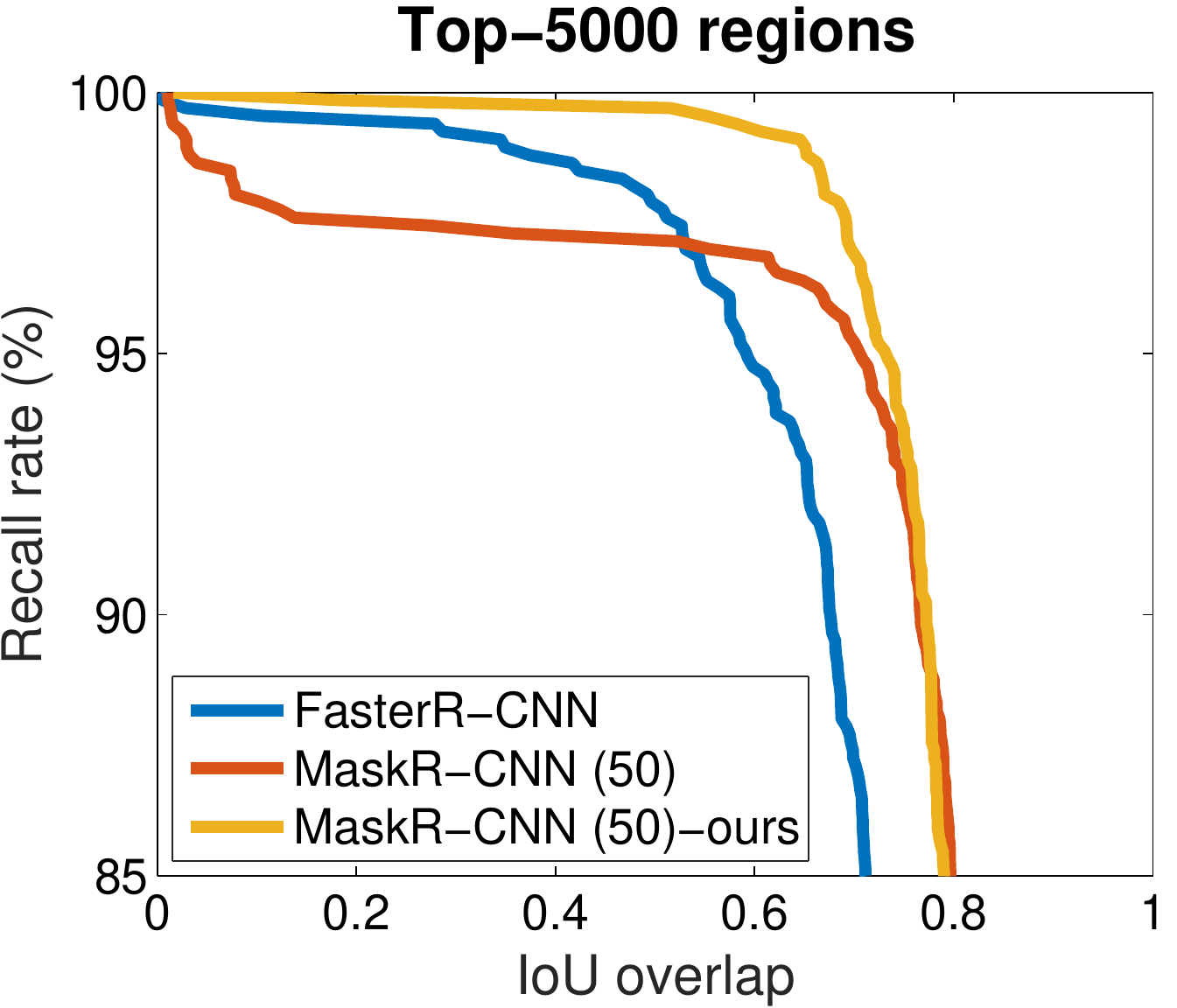}\label{fig:rpn-d}}
\caption{Miss rate and recall for region proposals generated by the RPN. Graphs (a) and (b) show results when considering all valid annotations, while graphs (c) and (d), when considering only groundtruth traffic signs in sizes of \si{30-50} pixels. We show in (a) and (c) miss rate over top-n regions using IoU overlap of 0.50, and in (b) and (d), recall rate over different IoU overlaps using the top 5000 region proposals. \label{fig:RPN-recall-all}}
\end{figure*}

The STSD benchmark contains around 20 categories with simple traffic signs in over \num{19236} images separated equally into the training (denoted \textit{Set1} in STSD) and the test set (denoted \textit{Set2}). However, only a subset of \num{3777} images from both sets contain annotations (denoted as \textit{Part0} in each set). We follow the evaluation protocol of~\cite{Zhu2016} and use only ten categories with images from \textit{Set1Part0} for the training and images from \textit{Set2Part0} for the testing. For fair evaluation with~\cite{Zhu2016}, we consider only annotations with bounding box sizes of at least \num{50} pixels. The remaining annotations are ignored in both the train and the test stage. Due to the GPU memory limitations, we resized images to have image size of at least 918 pixels (i.e., both width and height are at least 918 pixels). For fair comparison between different architectures, the same image size was used in all variants of Faster/Mask R-CNN. We did not use data augmentation in this experiment. 

Detailed results on STSD are reported in Tables~\ref{tab:STSD-details} and~\ref{tab:STSD-avg}, with the corresponding error rates in Figure~\ref{fig:STSD-graph}. When focusing on the related work and Faster/Mask R-CNN without our adaptations it is clear that pre-computed region proposals from R-CNN (as reported in~\cite{Zhu2016}) perform worse than the newer R-CNN variants with the region proposal network. Error rates for R-CNN are twice as large as for the Faster/Mask R-CNN. On the other hand, the fully convolutional method (FCN) proposed by~\cite{Zhu2016} achieves a significantly lower false-positive rate of \SI{2.3}{\%} than both Faster and Mask R-CNN, but has a slightly worse miss rate of \SI{7.1}{\%}. Faster and Mask R-CNN have a lower miss rate by \SI{1} percentage point (pp.). The standard mAP$^{50}$ metric in Table~\ref{tab:STSD-avg} also shows Faster R-CNN and Mask R-CNN with ResNet-50 achieving mAP$^{50}$ of \SI{94.3}{\%} and \SI{94.9}{\%}, respectively.

Results also show that the best performance is obtained when our adaptations are applied to the Mask R-CNN. Our proposed approach, in this case, achieves mAP$^{50}$ of \SI{95.2}{\%}, with average false-positive rate of \SI{2.5}{\%} and average miss rate of \SI{3.3}{\%}. Compared to the related work, the FCN~\cite{Zhu2016} achieves a similar false-positive rate but has at least twice as large miss rate at \SI{7.1}{\%}. Improvements in our approach are better reflected in F-measure, which is defined as a harmonic mean between precision and recall. Our approach clearly outperforms the state-of-the-art with \SI{2}{pp.} higher F-measure. Those improvements directly stem from our proposed adaptations and not from the Faster/Mask R-CNN as average miss and false-positive rates without our adaptations are still \SI{6.6}{\%} and \SI{4,7}{\%}, respectively, while they are reduced to only \SI{3.3}{\%} and \SI{2.5}{\%} with the proposed improvements. This is reflected in an improved F-measure and in mAP$^{50}$ as well.

\subsection{Evaluation on DFG traffic-sign dataset}

Next, the proposed method is evaluated on the DFG traffic-sign dataset. We use the train-test split as presented in Section~\ref{sec:dataset} with 200 categories in \num{5254} training and \num{1703} testing images, and using only annotations with at least 30 pixels in size. Annotations below 30 pixels are ignored during training and during evaluation we ignore detections of those objects to prevent penalizing the detector when it correctly detects small objects. We further resize images for both the training and the testing due to memory limitations. We resize images in all variants of Faster/Mask R-CNN to have image sizes of at least 840 pixels in both width and height. This was made for fair comparison under the same hardware limitations for all network models. Considering images are Full-HD with the image hight of 1080 pixels, this change represents slightly less than a \SI{25}{\%} reduction in size.

\paragraph*{Region proposal evaluation} We first evaluate the region proposal network separately from the classification network. This allows us to assess the quality of region proposals as generated by the RPN before they are passed to the classification module. We take top N regions from the RPN and observe miss rate and recall rate of all annotated traffic signs. To ensure correct balance between categories with either small or large number of instances, we calculate metric for individual categories and then report the average over all categories.

\begin{table}
\centering
\caption{Results on DFG traffic-sign dataset. \label{tab:DFG-map}}
\small{
\begin{tabular}{l c ccc }
\toprule
\multirow{3}{*}{} & \multirow{3}{*}{\makecell[l]{Faster\\R-CNN}} & \multicolumn{3}{l}{Mask R-CNN (ResNet-50)} \\
 \vspace{-8pt}\\
 \cline{3-5} 
\vspace{-5pt}\\
 & & No adapt. & \makecell[l]{With\\adapt.} &  \makecell[l]{With adapt. and\\data augment.}\\
\vspace{-7pt}\\
\midrule
mAP$^{50}$		& 92.4	& 93.0	&  95.2	& \textbf{95.5}	\\
mAP$^{50:95}$   & 80.4	& 82.3	&  82.0	& \textbf{84.4}	\\
Max recall     & 93.8	& 94.6	&  \textbf{96.5}	& \textbf{96.5}\\
\bottomrule
\end{tabular}
}
\end{table}
Results are reported in Figure~\ref{fig:RPN-recall-all}, with (a) - (b) showing results when all annotations are considered and with (c) - (d), for smaller traffic signs only, i.e., when considering only groundtruth traffic signs that are \si{30-50} pixels in size. In both cases, we report miss rate over the top-n regions using an IoU overlap of \num{0.50} in (a) and (c), and recall over different IoU overlaps using the top \num{5000} region proposals in (b) and (d). Figure~\ref{fig:rpn-b} first reveals that Faster R-CNN performs worse than the other methods. This is particularly evident at higher IoU overlaps where Faster R-CNN performs more than \SI{5}{pp.} worse.

\begin{table*}
\centering
\caption{Results on DFG traffic-sign dataset when considering different sizes of traffic signs. \label{tab:DFG-regions}}
\small{
\begin{tabular}{l lll lll llp{5pt} lll ll}
\toprule
\multirow{5}{*}{\makecell[l]{Traffic-sign size\\(\% signs retained)}} &  \multicolumn{2}{l}{Faster R-CNN} && \multicolumn{5}{l}{Mask R-CNN} && \multicolumn{5}{l}{\makecell[l]{Mask R-CNN with adapt.\\and data augmentation (ours)}}\\
 \vspace{-5pt}\\
 
\cline{2-3} \cline{5-9} \cline{11-15}
\vspace{-5pt}\\
 &&&& \multicolumn{2}{l}{ResNet-50} && \multicolumn{2}{l}{ResNet-101} && \multicolumn{2}{l}{ResNet-50} && \multicolumn{2}{l} {ResNet-101}\\
\vspace{-9pt}\\
 & \makecell[l]{Max\\recall} & mAP$^{50}$ && \makecell[l]{Max\\recall} & mAP$^{50}$ && \makecell[l]{Max\\recall} & mAP$^{50}$ &&  \makecell[l]{Max\\recall} & mAP$^{50}$ && \makecell[l]{Max\\recall} & mAP$^{50}$\\

\midrule
min 30 px (100\%) & 93.8	& 92.4	&& 94.6	& 93.0	&& 94.8	& 93.2	&& \textbf{96.5}	& \textbf{95.5}	&& 96.1	& 95.2\\
min 40 px (89\%) & 96.1	& 95.0	&& 96.8	& 95.3	&& 96.8	& 95.3	&& \textbf{97.4}	& \textbf{96.7}	&& 97.0	& 96.4\\
min 50 px (80\%) & 96.6	& 95.0	&& 96.7	& 94.9	&& 96.8	& 95.2	&& \textbf{97.2}	& \textbf{96.0}	&& 96.8	& 95.5\\

\bottomrule
\end{tabular}
}
\end{table*}

The miss rates of various top-n regions, shown in Figure~\ref{fig:rpn-a}, demonstrate that all methods perform extremely well with over \SI{99}{\%} of all traffic signs found. However, only our proposed method achieves close to zero miss rate, and as indicated by the recall over IoU overlaps in Figure~\ref{fig:rpn-b}, the proposed method is able to retain higher recall at higher overlap values. This suggests that our adaptations decrease the miss rate of the RPN and higher quality regions can be produced, i.e., regions with high overlap with the groundtruth. Moreover, improvements are more significant in smaller regions, as shown in Figure~\ref{fig:rpn-c} and~\ref{fig:rpn-d}. In this case, our adaptation achieves a significantly better miss rate than Faster/Mask R-CNN that did not use our adaptation. Even at a more liberal IoU overlap of \num{0.50}, the standard approach achieves a \SI{3}{\%} miss rate, while our adaptation achieves a miss rate close to zero. This difference is well observed in Figure~\ref{fig:rpn-d}, showing our proposed method achieving higher recall rates at larger IoU.

Improvements in the miss rate at this level are important for the whole pipeline, since objects missed by the region proposals at this stage cannot be recovered later by the classification network. Results show that Mask R-CNN is unable to achieve full detection of all objects, particularly for small objects; however, our adaptations overcome this issue and achieve a miss rate near zero. 

\paragraph*{Full pipeline evaluation} Next, we evaluate the whole detection pipeline with the RPN and classification networks combined. We report our results in terms of mean average precision (mAP) over all 200 categories as well as in terms of maximal possible recall that can be attained with the final detections when thresholding the score at \SI{0.01}. This value is directly related to the miss rate and the recall rate of region proposals in the previous section, and when both values are compared, we can deduce how many traffic signs were missed due to poor performance of the classification network only.

Results are reported in Table~\ref{tab:DFG-map} and clearly show that Faster R-CNN performs the worst among all methods, while the best results are achieved with our adaptations for Mask R-CNN. Nevertheless, all methods achieve mAP$^{50}$ of over \SI{90}{\%}. Compared to the original Mask R-CNN, our proposed adaptations already improve results when measured in mAP$^{50}$ and maximal recall/miss rate metrics, even without data augmentation. The performance in mAP$^{50}$ metric is improved from \SI{93}{\%} to over \SI{95}{\%}, and the miss rate error is almost halved from \SI{5.4}{\%} to \SI{3.5}{\%}. Slightly worse results are achieved in the mAP$^{50:95}$ metric but this is improved when augmentation is enabled. With augmentation we slightly improve mAP$^{50}$, and significantly improve mAP$^{50:95}$ from \si{82-83}{\%} with the original Mask R-CNN to \SI{84.4}{\%} for when our adaptations and data augmentation is used. Data augmentation has contributed mostly to improving the precision of bounding boxes. Results also reveal that while overall miss rate has been reduced by half compared to the original Mask R-CNN, there still remain \SI{3.5}{\%} missed objects despite, as shown in the previous section, having near zero miss rate in the region proposals. This points to traffic-sign detections being lost by the classification network.

\begin{figure}
\centering
\includegraphics[width=1.05\columnwidth]{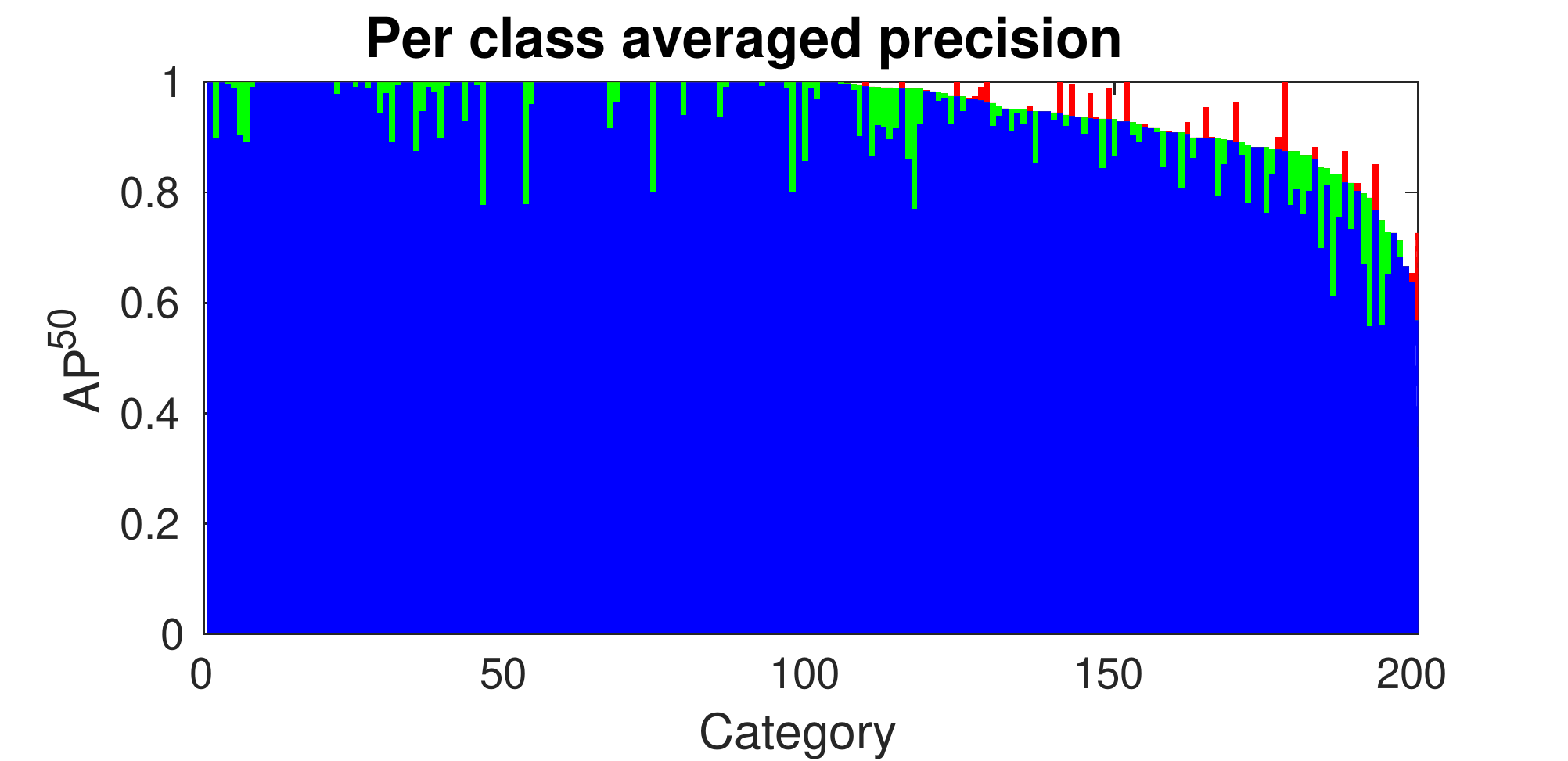}
\caption{Sorted per-class AP$^{50}$ distribution on the test set of the DFG traffic-sign dataset. The blue bars depict Mask R-CNN (ResNet-50) with our improvements and data augmentation, while green and red bars show change in performance (increased for green and decreased for red) compared to the base Mask R-CNN (ResNet-50) without our improvements.\label{fig:per-class-ap}}
\end{figure}

\begin{figure*}
\centering
\includegraphics[width=\textwidth]{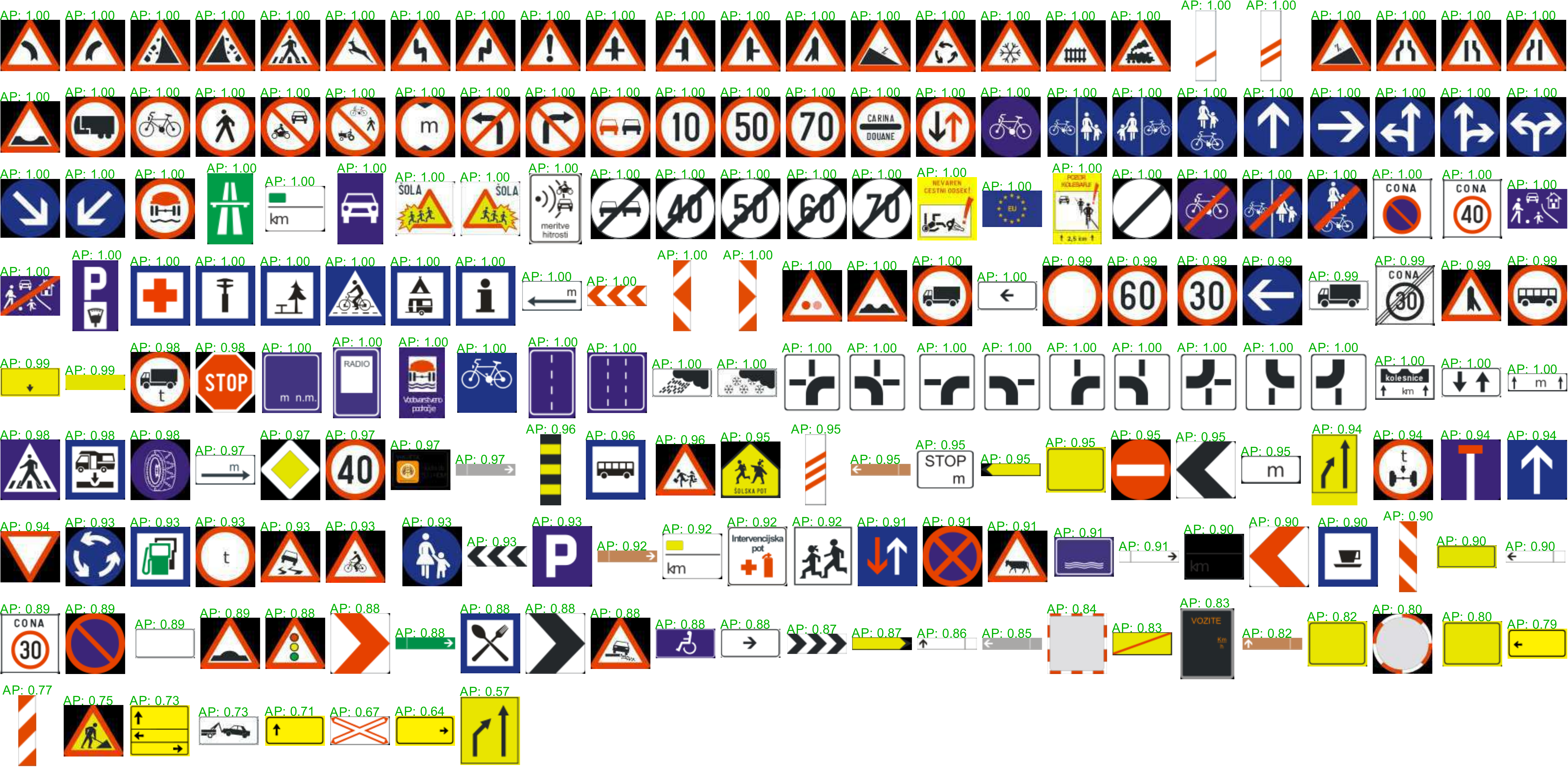}
\caption{DFG traffic-sign categories sorted by average precision (AP$^{50}$) calculated when using Mask R-CNN ResNet-50 with our adaptations and data augmentation. \label{fig:resnet-50-ours}}
\end{figure*}

\paragraph*{Different traffic-sign sizes}
We also perform evaluation considering different traffic-sign sizes with the results reported in Table~\ref{tab:DFG-regions}. This analysis reveals poor performance with smaller objects when using original Faster and Mask R-CNN. The difference in both mAP$^{50}$ and the maximal recall rate between small and large objects is around \SI{2}{pp.}. However, with our adaptations, the detection of smaller objects is improved significantly and completely eliminates the performance gap between detection of smaller and larger objects. Moreover, this is achieved on top of the improved detection for larger objects. 

\begin{figure*}[!b]
\includegraphics[width=\textwidth]{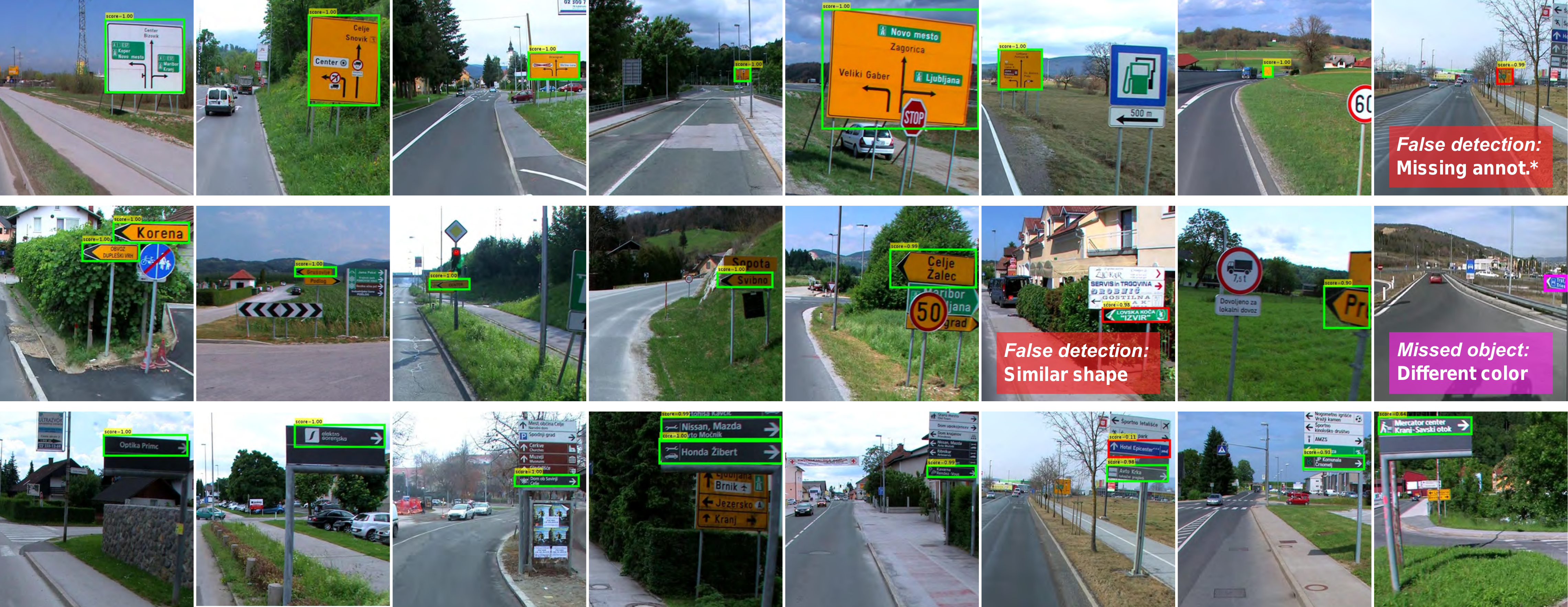}
\caption{Examples of complex traffic signs with variable content and good detection on the test set of the DFG traffic-sign dataset. True positives are depicted in green, false positives in red, and missing detections (false negatives) in magenta. (*) Note, the last detection in the first row is not false since actual traffic sign was not annotated due to high occlusion.\label{fig:examples-good-variable-app}}
\end{figure*}

\paragraph*{Deeper Residual Network}
We also show results with ResNet-101 architecture in Table~\ref{tab:DFG-regions}. ResNet-101 performs similarly to the smaller ResNet-50 in most cases. When our improvements are not included, ResNet-101 performs less than \SI{0.2}{pp.} better; however, this reverses when our improvements are included. The difference between both of them still remains minimal at below \SI{0.4}{pp.}. Since ResNet-101 is larger with twice as many number of layers with more computational resources required, the ResNet-50 represents a significantly better choice.

\section{Qualitative analysis}\label{sec:qualitative_analysis}

In this section, we demonstrate the performance of our approach on traffic-sign detection with additional qualitative analysis. We focus only on the best performing model, namely Mask R-CNN using ResNet-50 with our adaptations and data augmentation. All results in this section are reported on the test set of the DFG traffic-sign dataset.

A per-class distribution of AP$^{50}$ is depicted in Figure~\ref{fig:per-class-ap}. This graph clearly shows that a large number of traffic-sign classes (\num{108}) are detected and recognized with average precision of \SI{100}{\%}, i.e. with no errors. For the remaining categories our approach still achieved AP of above \SI{90}{\%} on 60 of them, and above \SI{80}{\%} on 23 of them.

\begin{figure*}
\includegraphics[width=\textwidth]{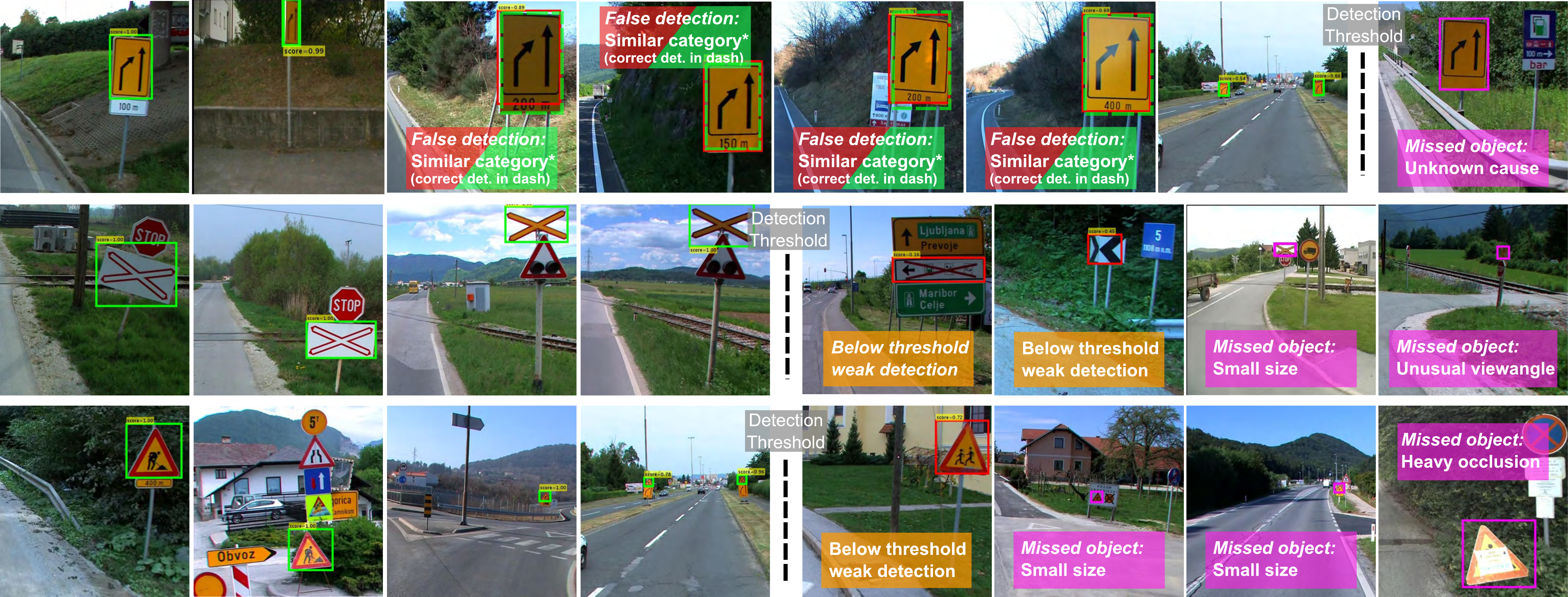}
\caption{Examples of traffic signs with fixed content but poor detection on the test set of the DFG traffic-sign dataset. True-positive detections are marked in green, false positives in red and missing detections (false negatives) in magenta. (*) Note that false detections in the first row occur due to two almost identical traffic-sign categories in the dataset (one with distance label below and one without). True detections with the other category detector are shown in dashed green line. \label{fig:examples-poor-fixed-var}}
\end{figure*}

Figure~\ref{fig:resnet-50-ours} further shows the traffic-sign classes with their corresponding AP$^{50}$ sorted by their AP$^{50}$ in descending order. The best performing categories at the top of the list are mostly traffic signs with low intra-category variations, i.e. with fixed sizes and fixed appearance. This includes various triangular danger signs, circular prohibitory signs, speed limit signs, rectangular information signs, etc. On the other hand, the worst performing signs at the bottom are traffic signs with a large variation of their sizes/aspect ratios as well as with a large intra-category variations, i.e., their content significantly varies from instance to instance. This includes particularly complex class of mirrors (both rectangular and round mirrors), speed feedback signs, various direction signs and signs marking the start or the end of the towns.

\paragraph*{Traffic signs with high intra-category variations and good performance} Figure~\ref{fig:resnet-50-ours} reveals several traffic signs with extremely good detection rate despite having large intra-category variations in their appearance. Samples for three such traffic-sign categories are depicted in Figure~\ref{fig:examples-good-variable-app}, namely they are: (i) \textit{large-direction-with-separate-lanes}, (ii) \textit{left-arrow-shaped-direction} and (iii) \textit{right-gray-direction}. Each row in this figure depicts one category with eight instances. For clarity we display only the relevant part of the image. True detections are shown in green, false detections in red and missing detections in magenta. Examples are also sorted by their descending detection score from left-to-right. Therefore if true (green) and false (red) positive detections can be successfully separated with a threshold then false detections can be trivially eliminated by setting an appropriate detection threshold. Note that this is important when looking at false detections as many of them are not problematic at all.

When focusing on the \textit{large-direction-with-separate-lanes} traffic-sign category in the first row in Figure~\ref{fig:examples-good-variable-app}, an extremely good performance is clearly shown for the traffic signs that have quite significant variation in their content as well as large variation in their sizes and aspect ratios. The first image in the top row depicts a good example of this as the traffic sign was detected with a high score despite having completely different color combination than other instances of the same class. Several detected instances are also quite small, yet our approach successfully detects them. Moreover, the last image in the first row shows a false detection of a small instance; however, a close inspection reveals that it is a correct detection. This instance was not annotated in the dataset due to small size and high occlusion of the tree.

The second row in Figure~\ref{fig:examples-good-variable-app} depicts detections of a \textit{left-arrow-shaped-direction} traffic sign. This category is fairly difficult to detect as aspect ratios vary quite significantly from instance to instance, mostly due to wide viewing angles, yet the detector did not have significant issues finding them. The second-to-last example in the second row is also significantly cropped; however, the detector is still able to correctly find it. 

Finally, detections for the \textit{right-gray-direction} traffic sign are shown in the last row in Figure~\ref{fig:examples-good-variable-app}. Detection of this category is difficult mostly due to significant variation of the content. Those traffic signs also often appear side-by-side in multiple rows which makes it difficult to generate the correct region proposal. Nevertheless, most instances have been correctly found.

\begin{figure*}
\includegraphics[width=\textwidth]{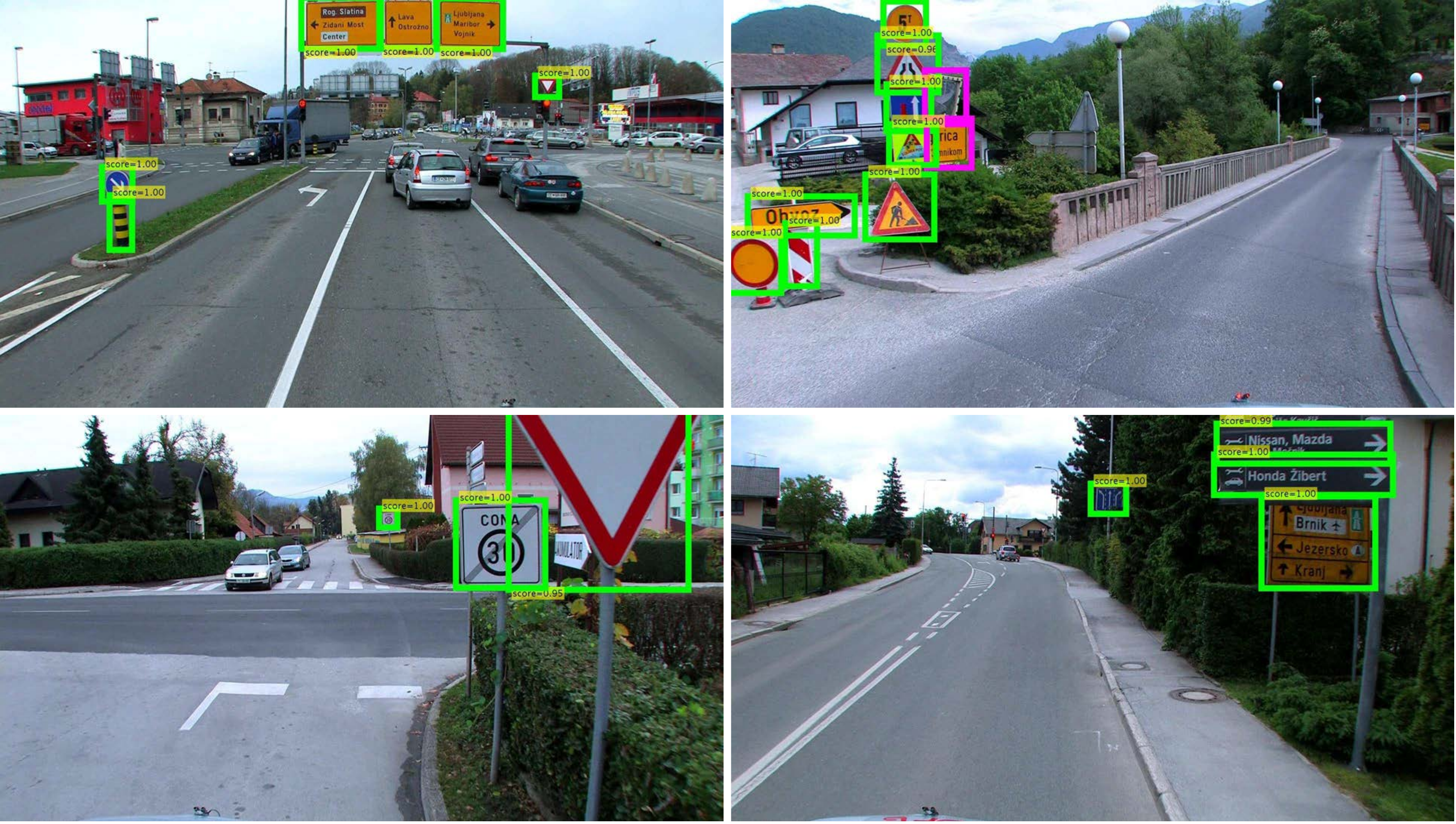}
\caption{Examples of detections on the test set of the DFG traffic-sign dataset. True detections shown in green and missing, in magenta.\label{fig:examples-all}}
\end{figure*}

\paragraph*{Traffic signs with poor performance and low intra-category variations} Next, we focus on three worst performing traffic signs despite having low appearance variation within a category, namely: (i) \textit{left-into-right-lane-merger}, (ii) \textit{train-crossing} and (iii) \textit{work-in-progress}. Samples are depicted in Figure~\ref{fig:examples-poor-fixed-var} and are organized in a similar manner as in Figure~\ref{fig:examples-good-variable-app}, with eight examples per category in a row, sorted by their descending detection score. 

The worst results are achieved for the \textit{left-into-right-lane-merger} traffic sign with the AP$^{50}$ of \SI{57}{\%}. Mask R-CNN correctly detects four out of five test instances, but appears to detect four false traffic signs as well, as can be seen in the top row. However, those false detections should not be considered problematic as the traffic sign is identical to the \textit{left-into-right-lane-merger} sign with the only difference in the distance value printed below the sign. Since the correct category is also detected (shown with the dashed green line), those false detections would be eliminated by the across-category non-maxima suppression, meaning that even in this case the issue is not as bad as it might seem. Still, such extremely minor differences between those two categories appear to pose a challenge for deep learning and point to a existing limitations of deep learning methods. 

The detector is also exhibiting inferior performance for the \textit{train-crossing} traffic sign as seen in the second row in Figure~\ref{fig:examples-poor-fixed-var}. The reason in this case can be found in two missed detections out of total six traffic signs. Both missed objects are very small, with one having fairly wide viewing angle, making the detection also extremely difficult. A few detections on false objects are also visible, most likely due to the presence of cross-like shape. However, they do not contribute to poor performance due to their low detection score.

The primary issue for the \textit{work-in-progress} sign, depicted in the third row of Figure~\ref{fig:examples-poor-fixed-var}, is high miss rate. Three out of eleven traffic signs are not detected. Most objects missed are also fairly small. The exception is the instance depicted in the last column where a significant occlusion would pose difficulty even for humans---its category was deduced from its inside color and the context. 

\paragraph*{Overall detection} 
Despite some missed detections shown in Figure~\ref{fig:examples-poor-fixed-var}, the detector still preforms extremely well even for several difficult cases. For instance, the second example in the first row of Figure~\ref{fig:examples-poor-fixed-var} is extremely difficult to detect due to a large viewing angle, but the detector still managed to find it---even with a large score. The detector was also able to find some fairly small instances, such as ones in the first and the last row.

Good performance is also reflected in Figure~\ref{fig:examples-all} where all traffic-sign detections are displayed for a couple of full-resolution images. This figure shows detections of several complex instances with occlusions and small traffic sign sizes; however, the detector still performs extremely well. 

\section{Discussion and conclusion} \label{sec:conclusion}

In this work, we have addressed the problem of detecting and recognizing a large number of traffic-sign categories for the main purpose of automating traffic-sign inventory management. Due to a large number of categories with small inter-class but high intra-class variability, we proposed detection and recognition utilizing an approach based on the Mask R-CNN~\cite{He2017} detector. The system provides an efficient deep network for learning a large number of categories with an efficient and fast detection. We proposed several adaptations to Mask R-CNN that improve the learning capability on the domain of traffic signs. Furthermore, we proposed a novel data augmentation technique based on the distribution of geometric and appearance distortions. As an important contribution, we also present a novel dataset, termed the DFG traffic-sign dataset, with a large number of traffic-sign categories that have low inter-class and high intra-class variability. This dataset has been made publicly available together with the code for our improvements, allowing the research community to make further progress on this problem and enabling reliable and fair comparison of different methods on a large-scale traffic-sign detection problem. We also  extensively evaluated our proposed improvements and compared them against the original Faster and Mask R-CNN. Our evaluation on the DFG and the Swedish traffic-sign datasets showed that the proposed adaptations improve the performance of Mask R-CNN in several metrics. This includes improvement in the miss rate of the RPN network for smaller objects, improvement in the overall recall of the full pipeline for both small and large objects, as well as improvement in the overall performance in the mean average precision. 

Our qualitative analysis further revealed how a \si{2-3}{\%} average error rate is reflected in actual detections. This is well demonstrated in Figure~\ref{fig:examples-all} where detections of several complex traffic-sign categories are depicted. Overall, we showed that the deep learning based approach is able to achieve extremely good performance for many traffic-sign categories, including several complex ones with large intra-class variability. Large error rates for problematic traffic-sign categories are mostly due to similarity to other categories, wide viewing angles and large occlusions. However, those issues do not pose a problem for the application of maintaining an accurate record of traffic-sign inventory. They can be mitigated by the detection over several video frames or matching 3D locations from stereo cameras. In particular, this system is already being deployed for traffic-sign inventory management on Slovenian roads. However, the proposed solution is also applicable to other problems requiring the capability of traffic-sign detection such as autonomous driving and advanced driver-assistance systems.

Despite excellent performance of the proposed approach there is still room for improvement. Our analysis revealed that the ideal performance is still not achieved, mostly due to several missed detections that are being lost by the classification network. Future improvements should focus on improving this part of the system.

 \section*{Acknowledgements}
 \small{
 This work was in part supported by the ARRS research project L2-6765 (ViLLarD) and ARRS research programme P2-0214. We would also like to thank the company DFG Consulting d.o.o., in particular Domen Smole, Simon Jud and mag. Toma\v{z} Gvozdanovi\'{c}, for capturing and annotating images and for their help in creating the dataset.}

{\small
\bibliographystyle{IEEEtran}
\bibliography{library}
}
\newpage
\vspace{-130px}
\begin{IEEEbiography}[{\includegraphics[width=1in,height=1.25in,clip,keepaspectratio]{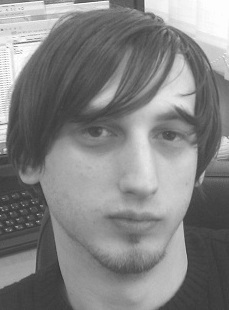}}]{Domen Tabernik} received a bachelors degree in Computer and Information Science in 2010. From 2010 he has been working as a computer vision researcher in the Visual Cognitive System Laboratory of Faculty of Computer and Information Science in University of Ljubljana. Since 2015 he is enrolled in the doctoral program of Faculty of Computer and Information Science in University of Ljubljana where he is working on the topics of compositional hierarchies and deep learning.
\end{IEEEbiography}
\vspace{-530px}

\begin{IEEEbiography}
[{\includegraphics[width=1in,height=1.25in,clip,keepaspectratio]{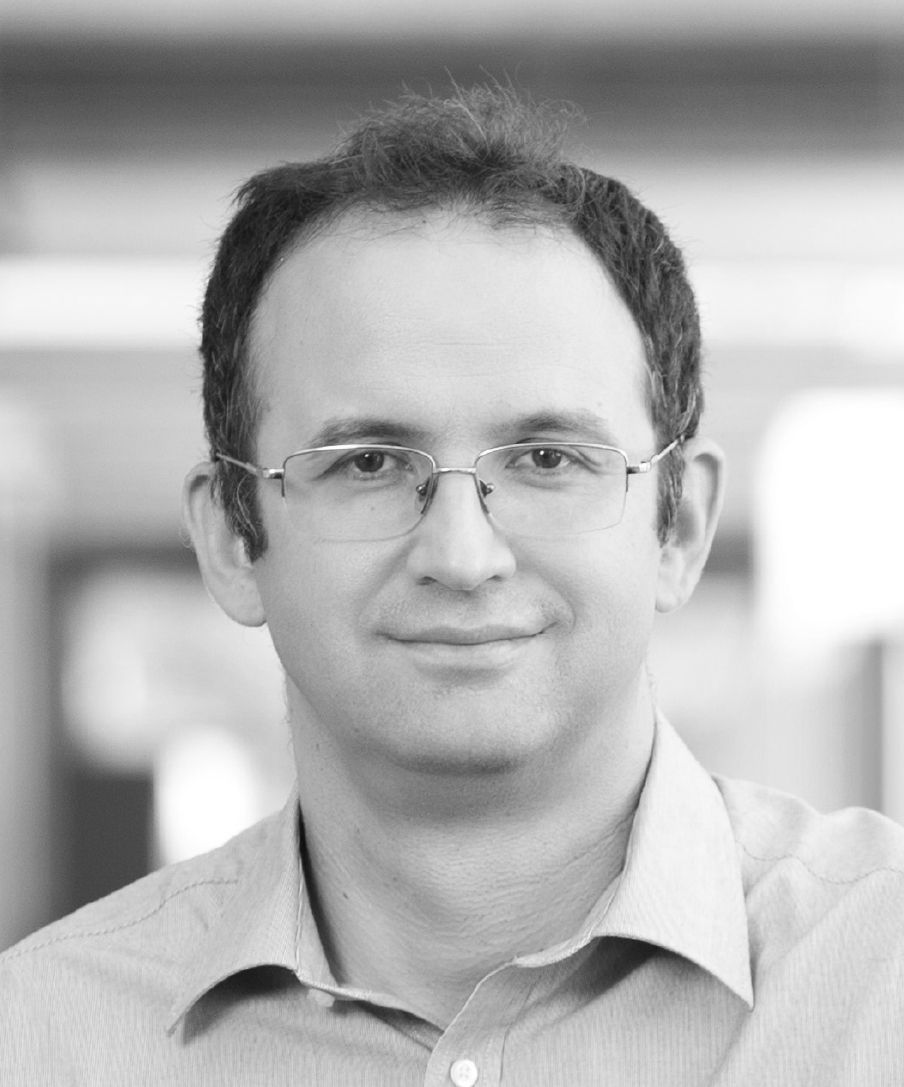}}]{Danijel Sko\v{c}aj}
is an associate professor at the University of Ljubljana, Faculty of Computer and Information Science. He is the head of the Visual Cognitive Systems Laboratory. He obtained the Ph.D. in computer and information science from the University of Ljubljana in 2003. His main research interests lie in the fields of computer vision, pattern recognition, machine learning, and cognitive robotics. 
\end{IEEEbiography}
\end{document}